\documentclass[journal]{IEEEtran}
\pdfoutput=1
\usepackage{titling}

\usepackage{ifpdf}
\usepackage{cite}
\usepackage{colortbl}
\usepackage{xcolor}
\newcommand\crule[3][black]{\textcolor{#1}{\rule{#2}{#3}}}

\usepackage{hyperref}
\definecolor{A1}{RGB}{140,67,46}
\definecolor{A2}{RGB}{0,0,255}
\definecolor{A3}{RGB}{255,100,0}
\definecolor{A4}{RGB}{0,255,123}

\definecolor{A5}{RGB}{164,75,155}
\definecolor{A6}{RGB}{101,174,255}
\definecolor{A7}{RGB}{118,254,172}
\definecolor{A8}{RGB}{60,91,112}

\definecolor{A9}{RGB}{255,255,0}
\definecolor{A10}{RGB}{255,255,125}
\definecolor{A11}{RGB}{255,0,255}
\definecolor{A12}{RGB}{100,0,255}

\definecolor{A13}{RGB}{0,172,254}
\definecolor{A14}{RGB}{0,255,0}
\definecolor{A15}{RGB}{171,175,80}
\definecolor{A16}{RGB}{101,193,60}

\definecolor{B1}{RGB}{192,192,192}
\definecolor{B2}{RGB}{0,255,0}
\definecolor{B3}{RGB}{0,255,255}
\definecolor{B4}{RGB}{0,128,0}
\definecolor{B5}{RGB}{255,0,255}
\definecolor{B6}{RGB}{165,82,41}
\definecolor{B7}{RGB}{128,0,128}
\definecolor{B8}{RGB}{255,0,0}
\definecolor{B9}{RGB}{255,255,0}

\usepackage{verbatim}
\ifCLASSINFOpdf
\usepackage[pdftex]{graphicx}
\DeclareGraphicsExtensions{.pdf,.jpeg,.png}
\usepackage[cmex10]{amsmath}
\usepackage{algorithmic}
\usepackage{amsfonts}
\usepackage{amssymb}

\newtheorem{theorem}{Definition}

\graphicspath{{Figures/},{SuperpixelFigures/}}

\usepackage{array}

\usepackage[caption=false,font=normalsize,labelfont=sf,textfont=sf]{subfig}

\usepackage{fixltx2e}

\begin{document}
\title{\Huge Superpixel Contracted Graph-Based Learning for Hyperspectral 
Image Classification}

\author{Philip~Sellars$^1$, Angelica I. Aviles-Rivero$^2$ and Carola-Bibiane 
Sch{\"o}nlieb$^1$ \thanks{P. Sellars and Carola-Bibiane Sch{\"o}nlieb are with 
the Department of Theoretical Physics and Applied Mathematics, Univeristy of 
Cambridge, Cambridge, UK.  {ps644,cbs31}@cam.ac.uk. AI Aviles-Rivero is with 
the Department of Pure Mathematics and Mathematical Statistics (DPMMS), 
Univeristy of Cambridge, Cambridge, UK. ai323@cam.ac.uk}}

%
%

\markboth{Journal of \LaTeX\ Class Files,~Vol.~13, No.~9, September~2014}%
{Shell \MakeLowercase{\textit{et al.}}: Bare Demo of IEEEtran.cls for Journals}
%


\maketitle

\begin{abstract}
A central problem in hyperspectral image classification is obtaining high classification accuracy when using a \textit{limited amount of labelled data}. In this paper we present a novel graph-based framework, which aims to tackle this problem in the presence of large scale data input. Our approach utilises a novel superpixel method, specifically designed for hyperspectral data, to define meaningful local regions in an image, which with high probability share the same classification label. We then extract spectral and spatial features from these regions and use these to produce a contracted weighted graph-representation, where each node represents a region rather than a pixel. Our graph is then fed into a graph-based semi-supervised classifier which gives the final classification. We show that using superpixels in a graph representation is an effective tool for speeding up graphical classifiers applied to hyperspectral images. We demonstrate through exhaustive quantitative and qualitative results that our proposed method produces accurate classifications when an incredibly small amount of labelled data is used. We show that our approach mitigates the major drawbacks of existing approaches, resulting in our approach outperforming several comparative state-of-the-art techniques.
\end{abstract}


%
\IEEEpeerreviewmaketitle

\section{Introduction}
\IEEEPARstart{I}N modern applications, hyperspectral images (HSI) capture a detailed light distribution, over several hundreds of spectral bands. This detailed spectral and spatial information increases the discriminative ability of HSIs compared to conventional colour images or multi-spectral images. As a result, hyperspectral imaging has been used in a wide range of applications including classification~\cite{melgani2004classification,fang2015spectral,fang2018new}, object tracking~\cite{wang2010bio,uzkent2016real,uzkent2017aerial}, environmental monitoring~\cite{ellis2004evaluation, manfreda2018use} and object detection~\cite{pan2003face, liu2017tensor, zhang2017joint}. 

In recent years, the classification of hyperspectral data has been an active topic of research. Classifying HSIs requires assigning a class label to each pixel within the image. There are several large hurdles to overcome during the classification process: the high dimensionality of the spectral information, the large spatial variability of the data and the limited number of training samples available due to the cost of data labelling. There have been numerous different attempts to deal with these problems when classifying HSIs, in which the majority of solutions rely on supervised learning (SL).

Kernel based classifiers such as support vector machines (SVM) \cite{melgani2004classification,SVMOld} are commonly used in the field. Whilst initial kernel methods only used spectral features, many later kernel methods included spatial features. An example being the multiple kernel learning (MKL) aproach of Fang et al \cite{SCMK} which used MKL to combine spatial based feature vectors alongside spectral features.

\begin{figure}
	\centering
    \includegraphics[width=0.5\textwidth,height=65mm]{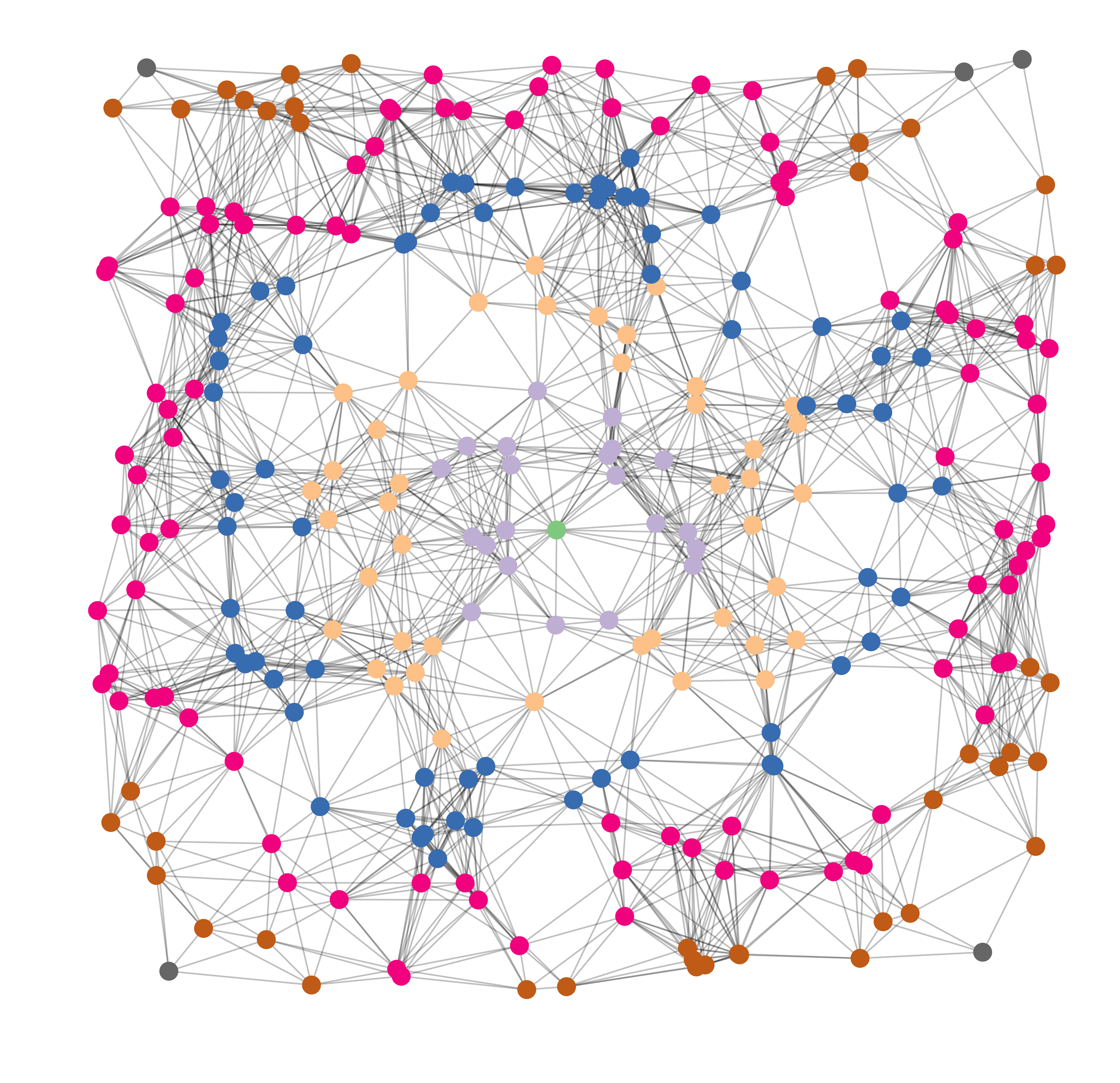}
	\caption{Data visualisation using a weighted undirected graph. The different node colours represents the minimum path length between each node and the central node of the graph, which is coloured in green. Graphs are incredibly useful tools for capturing and visualising the detailed information present in data. Furthermore, graphs are particularly useful for visualising high dimensional data such as that present in hyperspectral images.}
	\label{fig::teaser}
\end{figure}

To deal with the high dimensionality of the data, many different feature extraction (FE) methods have been investigated. These methods aim at finding a lower dimensional subspace in which the separability among samples is maximised. Kang et al used image fusion and recursive filtering to extract meaningful features \cite{IFRF}, Li at al \cite{LBP} exploited local binary patterns to extract local features and textural information and Fang et al \cite{LCMR} used local co-variance matrix representation to characterize the correlation between the spectral and spatial information in HSI data.  

Motivated by the remarkable success of Deep Learning (DL), different works have used DL for HSI classification. Convolutions neural networks (CNN) are commonly used to extract high level spectral and spatial features \cite{CNN1} \cite{CNN2}. Makantasis et al \cite{CNN1} used a CNN to extract spatial and spectral features and passed these into a multi-layer perceptron. In recent work, generative adversarial networks (GAN), which simultaneously train a generator and discriminator, have also been explored for HSI classification \cite{HGAN}.  

Although SL based classifiers have shown good results on HSI data, their performance is heavily reliant on having a large quality training set which is a costly investment. As an alternative to SL, we could use unsupervised learning (UL), in which the key idea is to rely on learning a set of classes from data that has not been labelled\cite{UL}. Although works such as \cite{ULHB} reported promising results on using UL for HSI classification, the major problem with UL is that the classification task becomes a massively ill-posed problem that needs specific assumptions to mitigate the lack of correspondence between the produced clusters and the known classes.

The aforementioned constraints associated with SL and UL make semi-supervised learning (SSL)\cite{SSTheory} a clear alternative for obtaining an improved classification performance. The idea of SSL is to exploit both labelled and unlabelled data in the training process to produce a higher classification accuracy than solely using the labelled data. The advantages of SSL when using HSI data are two-fold: we decrease the need for large amounts of labelled data and we gain further understanding of the relationships present in the data.  

In this paper, we introduce for the first time a superpixel contracted graph-based learning framework for  semi-supervised HSI classification, that we named \textit{Superpixel Graph Learning} (SGL). It produces state-of-the-art results, especially when the amount of labelled data is \textit{extremely} small. Our framework is composed of three main parts. Firstly, we use a novel superpixel algorithm, specially designed for HSIs, to accurately partition our images into adaptive regions termed superpixels. Secondly, we perform feature extraction on each superpixel to extract discriminative features. Finally, we use the superpixels and features to produce a weighted graphical representation of our image which is then classified using a graphical-learning method (LGC \cite{LGC}). Our main contributions are:

\begin{itemize}
\item We propose a novel computationally tractable framework for HSI classification, in which our novelty largely relies on:
    \begin{itemize}
        \item \textbf{A hyperspectral superpixel approach}. To the best of our knowledge, this is the first time that a superpixel approach has been designed specifically for HSI data, that is an approach which considers both spatial and spectral information. We proposed a new novel clustering distance, which combines a Euclidean spectral distance with the Log-Euclidean distance of a covariance matrix representation. This allows us to define meaningful local regions to boost the overall classification performance.
        
        \item  \textbf{Superpixel graph classification}. We show that combining superpixels with a graphical representation and a purely graphical classifier brings two major advantages: firstly, it vastly decreases the size of the node set which allows for classification in computationally feasible times without the need for matrix approximation methods. Secondly, it allows for the intelligent regularisation of the final classification map by using superpixels as adaptive local regions.
    \end{itemize}
    
  \item We extensively validate our proposed approach by using three benchmarking datasets and provide a range of experimental results. 
  \item We demonstrate that the combination of our novel hyperspectral superpixel approach embedded in a graphical setting leads to state of the art results for HSI classification. 
\end{itemize}

The remainder of this paper is organised as follows. Section II explores the related work on semi-supervised learning in the context of HSI classification. Section III is devoted to describing the proposed SGL method including superpixel generation, feature extraction and graph-based semi supervised classification. Section IV contains the experimental results for testing upon three real HSIs and a comparison to other state-of-the-art classification methods. Finally, Section V  presents the conclusions as well as discussion of further work.

\begin{figure*}[t!]
    \centering
    \includegraphics[width=\textwidth]{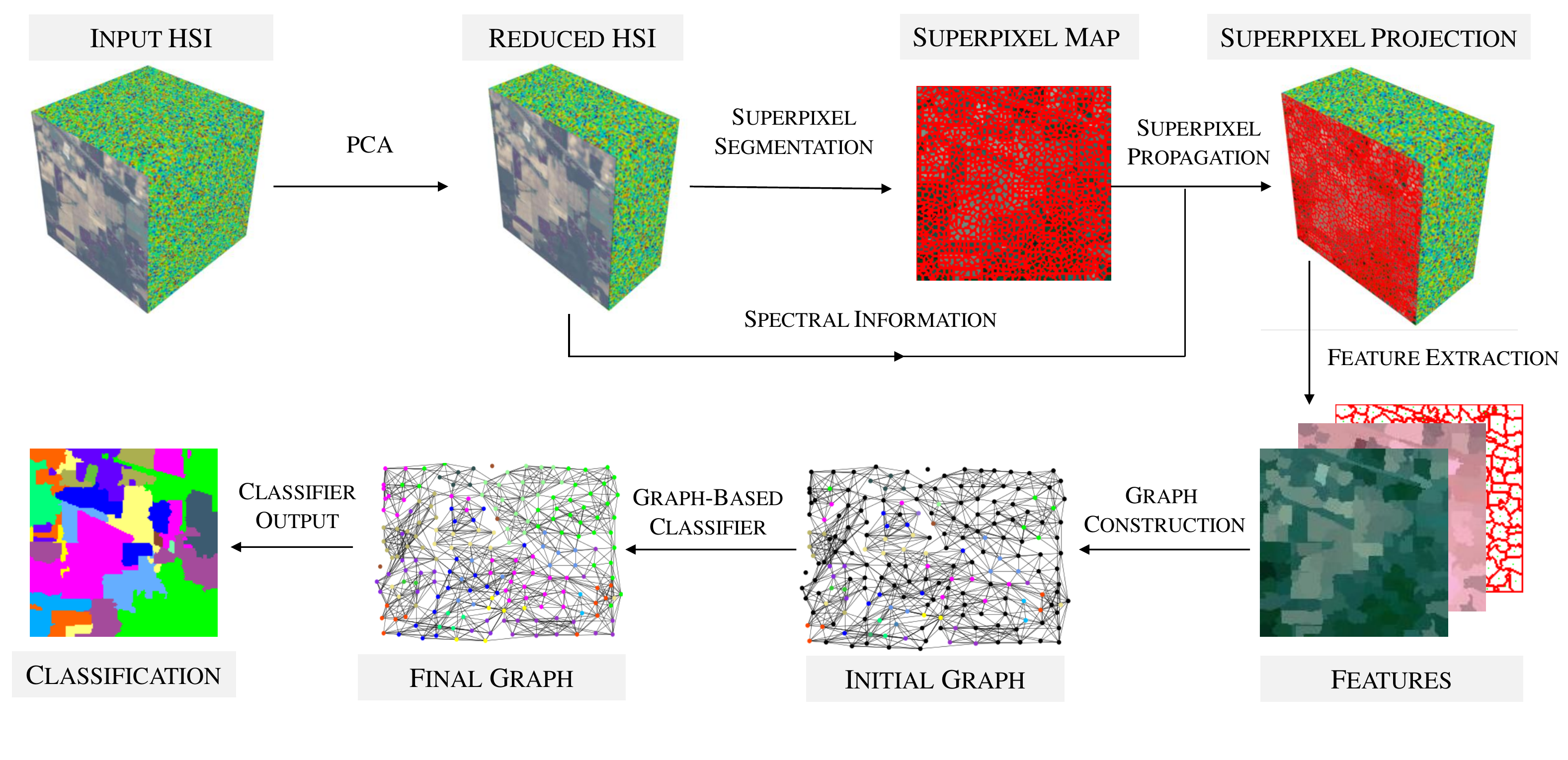}
    \caption{The proposed framework for the method. A HSI is read in and dimensionally reduced before superpixel segmentation occurs. Features are then extracted from each superpixel and, when combined with the initial labelling, are used to create a superpixel based graph. A graph classifier is used to propagate information across the graph. The final labels are then combined with the superpixel map to give the classification of the HSI.}
    \label{fig:workflow}
\end{figure*}

\section{Related Work}
The problem of semi-supervised classification of HSIs has been previously investigated by the remote sensing community. In this section, we review the existing techniques in turn. The literature regarding Semi-Supervised Learning (SSL) algorithms can be roughly categorised into three different categories. These being \textit{generative models}, \textit{low-density separation} and \textit{graph-based methods}.

Several previous methods have utilised \textit{graph-based learning}, and our paper is closely related to these. Graph-based methods rely upon constructing a graph representation, where the data points are represented by nodes and the similarity between these data points shown by edges and weights (see Fig \ref{fig::teaser}).
The first graph-based learning method was proposed by Camps-Valls in~\cite{Hgraph}. This paper used different spectral and spatial kernels alongside the Nystr{\"o}m extension, as a matrix approximation tool, to classify HSIs in computationally reasonable times. However, the produced accuracy was poor compared to other methods at the time. Gao et al \cite{bilinearGraph} used a bilayer graph-based learning algorithm to improve classification performance. The two layers were composed of a pixel-based graph, similar to \cite{Hgraph}, and a hypergraph built from grouping relations estimated using unsupervised learning. Cui et al \cite{SERW} used an extended random walker (ERW) on a superpixel-based graph to optimise a classification map produced from an SVM. Showing that the accuracy of the SVM could be greatly improved by using the information present in the graph.  

Another group of semi-supervised methods seek to directly implement the \textit{low density separation} assumption \cite{SSTheory} by moving the decision boundary away from unlabeled points. The first paper published in this area was by Bruzzone et al \cite{TSVM} which used a novel transductive SVM (TSVM) for HSI classification. A TSVM differs from the standard SVM as it seeks to maximise the margin on a combination of labeled and unlabeled data. Building upon these ideas came semi-supervised self-learning algorithms such as the work by Dópido et al \cite{selfL}, in which they sought to adapt active learning, in a which a user actively selects unlabeled samples, to a self learning framework in which the computer automatically selects the most informative unlabeled samples for classification purposes. Ratle et al \cite{SSNN} took a different path and tackled low density separation using a semi-supervised neural network architecture. An embedding regularizer was added to the loss function to inject the unlabeled information and this approach produced higher classification accuracy than TSVMs.

The rise of deep learning methods, has led to an increase in popularity of \textit{generative methods} for semi-supervised learning. However, these methods are in still in their infancy. One of the most popular approaches by Zhan et al \cite{HGAN2} uses a generative adversarial network (GAN) to simultaneous train a discriminator and generator. However, this paper uses a 1D-GAN and can only exploit spectral feature and the produced accuracy suffers as a result. Zhu et al \cite{HGAN} developed a 3D-GAN which used convolutional neural networks for the discriminator and generator. This architecture allows the approach to exploit the spectral-spatial information present in the HSI. Therefore, the produced accuracy was much higher than \cite{HGAN2}.

Although works based on generative models and low-density separation have shown encouraging results, in this work, we concentrate on producing a graph-based method, the motivation for which is three-fold. Firstly, data can be naturally represented on graphs. Secondly, a graph representation is motivated by its mathematical background and properties including spareness. Thirdly, data can be represented in an uniform space even if the data is highly heterogeneous. We seek to produce a graph-based method that is based on superpixel representations similar to that of \cite{SERW}. However, unlike \cite{SERW} we seek to produce a fully graph-based learning method rather than a graph-based optimisation of a non-graph based method.

\section{Proposed Method}
This section is devoted to explaining our proposed framework, which we call SGL. It contains three main parts which are shown in Fig \ref{fig:workflow}. Firstly we describe our hyperspectral superpixel algorithm, subsequently we give a description of the feature extraction process and finally we describe the construction and classification of our graph representation. \smallskip

\textbf{Problem Statement.} In this work, we seek to find an accurate classification prediction for a large amount of unlabelled data given an extremely small amount of labelled data. We consider the following problem definition for the classification task under the SSL paradigm.

\begin{theorem}
    \textbf{Semi-supervised Classification Task.} Given a set of points  $\{ (x_i ,y_i) \}_{i=1}^{l}$,  $\{ x_k \}_{k=l+1}^{l+u}$,
    and a label set  $\mathcal{L} = \{1,..,c\}$ where $\{y_i\}_{i=1}^{l} \in \mathcal{L}$, then, 
    we seek to find a function $f: \mathcal{X}^{l+u} \mapsto \mathcal{Y}^{l+u}$, which utilises the unlabelled data $\{ x_k \}_{k=l+1}^{l+u}$, such that $f$ allows for a good prediction for $\{ x_k \}_{k=l+1}^{l+u}$.
\end{theorem}

\subsection{Superpixel Segmentation}
\begin{figure}[t!]
    \centering
    \includegraphics[width=0.5\textwidth]{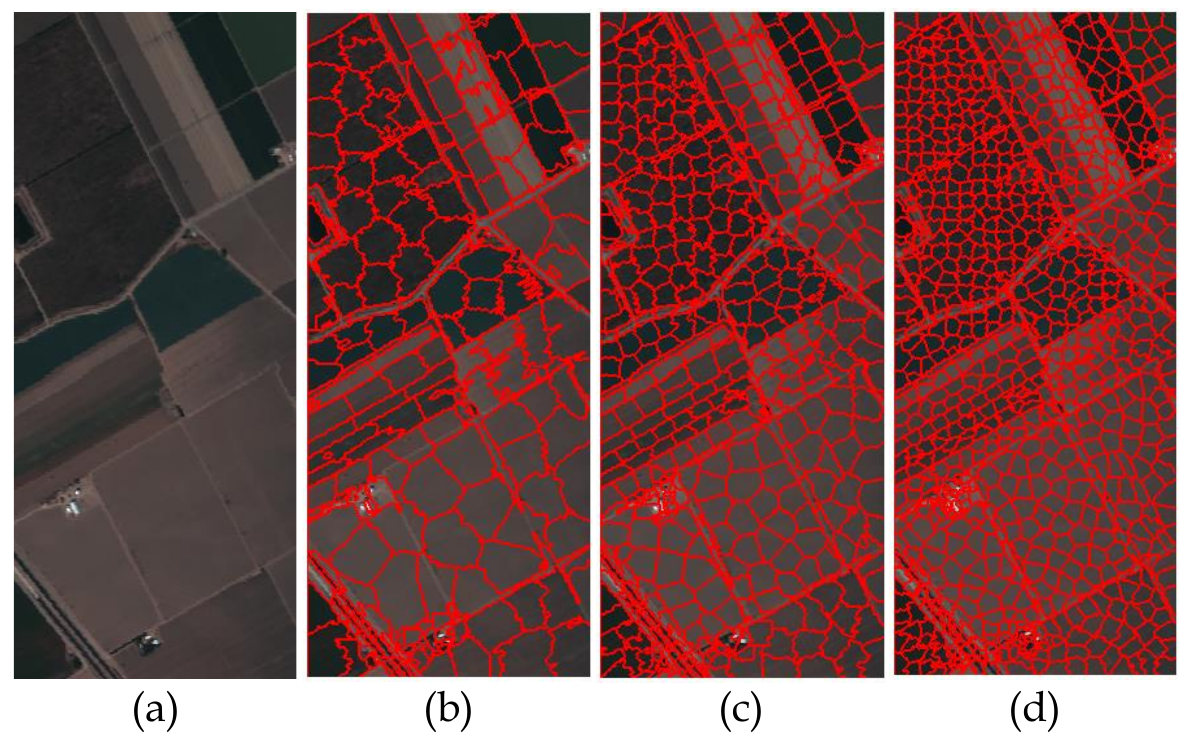}
    \caption{The Salinas HSI segmented using our proposed HMS algorithm. Fig (a) shows a RGB version of the image and Figs (b)-(d) show the image segmented using 280, 569 and 1034 superpixels respectively. Note that due to content sensitive nature of the HMS extension, there are a larger number of smaller superpixels in content dense regions.}
    \label{fig:salinas}
\end{figure}

Superpixels are perceptually meaningful connected regions which group pixels similar in colour or other features and were initially introduced by Ren and Malik \cite{superpixels}. In subsequent years, many different algorithmic approaches have been proposed (e.g.\cite{slic,MSLIC,2018arXiv180202796M}). For a detailed survey on superpixel algorithms see~\cite{survey}. Fig \ref{fig:salinas} shows the application of a superpixel algorithm to a HSI. Superpixel maps such as the one shown in Fig \ref{fig:salinas} have many desirable properties: they are computationally and representationally efficient, the individual superpixels are perceptually meaningful and as superpixels are the result of an over-segmentation they are very good at conserving image structures.

\textit{Why use superpixels as a tool for HSIs?} In order to extract spatial features for use in spectral-spatial models, it is important to be able to define good local regions. Whilst setting a fixed size window (e.g.~\cite{chen2011hyperspectral}) has shown good results, a fixed size does not allow for the full exploitation of spatial context. Using superpixels as adaptive regions \cite{SCMK} has been shown to produce  discriminative information. Cui et al.~\cite{SERW} demonstrated this by using a superpixel based random walker to optimise an SVM probability map to great effect. Furthermore, Cui et al. additionally demonstrated that a superpixel spectrum is more stable and less affected by noise that an individual pixel spectrum. Therefore by using superpixels we become more resistant to noise present in the data. 

The most common algorithm used in clustering based superpixel methods is Lloyd's algorithm~\cite{lloyd1982least}, a modified version of the popular k-means clustering algorithm. In the context of Lloyd's algorithm, let us first formalise the definition of a superpixel segmentation.
\begin{theorem}{\textbf{Superpixel Segmentation.}} 
 Given an image $I:\mathcal{X}\rightarrow \Omega$, where $\mathcal{X} \subset \mathbb{Z}^2$, a superpixel over-segmentation is a partition  $\{\mathcal{S}_i\}_{i=1}^n$ of $\mathcal{X}$  such that for each $1\leq i\leq n$ we have $S_i=\{x:d((x,I(x)),F(S_i))$, where $d$ is a metric, $F$ is a feature function and $\mathcal{S}_i$ is an individual superpixel. 
\end{theorem}
\smallskip

In this work, we build on this definition to propose our algorithmic approach. Denote a HSI as $\textbf{I} = \{I_{b}\}, b= 1,..,\mathcal{B}$ with dimensions $\mathcal{W} \times \mathcal{H} \times \mathcal{B}$ representing the width,  height and number of bands respectively.  Firstly, for computational efficiency we use PCA \cite{PCA} to reduce the dimensionality and produce a reduced image $\mathbf{\hat{I}} = \{\hat{I}_{a}\}, a= 1,..,\mathcal{A}$  where $\mathcal{A} \ll \mathcal{B}$.  Denoting an individual pixel as $p \in \mathbf{\hat{I}}$, we then seek to partition our reduced HSI $\mathbf{\hat{I}}$ into superpixels. This corresponds to splitting $\mathbf{\hat{I}}$ into a family of disjoint sets, $ \mathbf{\hat{I}} = \cup_{i=1}^{K} \mathcal{S}_i$ , $\mathcal{S}_i \cap \mathcal{S}_j = \emptyset $, where  $\mathcal{S}_i$ corresponds to an individual superpixel and $K$ is the number of superpixels. Each superpixel $\mathcal{S}_i$ is made up of a set of $n_i$ connected pixels, $\mathcal{S}_i =  \{ p_{i,1} , ... , p_{i,n_i}  \}$.

\textit{Hyperspectral superpixel construction.} When constructing our hyperspectral superpixels, we need to ensure that our algorithm extracts effective information from hyperspectral data. Whilst other works, including~\cite{SCMK}, feed the first three principal components of HSIs into RGB based superpixel algorithms, we seek to design an algorithm specifically built for hyperspectral data to ensure good performance.

As the base for our algorithm, we began with Manifold SLIC (MSLIC) \cite{MSLIC}. MSLIC has two features that make it highly useful for our purpose. Firstly it produces content sensitive superpixels by mapping the image $I$ to a two dimensional manifold $\mathcal{M}$ and measuring the area of Voronoi cells on $\mathcal{M}$. Secondly, the number of superpixels will change from the initial selection to fit the content structure in the image, thereby lowering the chances of a poor initial choice of $K$ greatly reducing the final accuracy. 

Our proposed method is an novel extension of MSLIC into hyperspectral data. We name this extension Hyper Manifold SLIC (HMS). HMS involves three major changes over MSLIC.

\noindent \subsubsection{High dimensional adaption} We alter the MSLIC 
algorithm to take image data with any number of bands $\mathcal{B}$. This 
involves changing several steps such as mapping the image $\mathbf{{I}}$ to a 
2-dimensional manifold $\mathcal{M} \in \mathbb{R}^{\mathcal{B}+2}$ rather than 
the standard $\mathcal{M} \in \mathbb{R}^{5}$. 

\noindent \subsubsection{Hyperspectral clustering distance} Based on our previous work \cite{OldPaper}, we design a more effective clustering distance as a combination of the Euclidean spectral distance~\cite{2018arXiv180202796M} and Log-Euclidean (LED) distance~\cite{mbdm} of a covariance matrix representation~\cite{CMR}. This combination effectively combines the spatial and spectral data present in the image. For each pixel ${p} \in \mathbf{\hat{I}}$ we construct a covariance matrix $\textbf{C}_{{p}}$ using the same methodology as Fang et al~\cite{LCMR} and use the LED metric to calculate the distances between these matrices. The distance between two pixels $p_x$, $p_y$ is given by:

\begin{equation}
\begin{aligned}
    d(p_x,p_y) =  || \text{logm} (\textbf{C}_{p_x}) - \text{logm} (\textbf{C}_{p_y}) ||_F \\
    + || \hat{\mathbf{I}}(p_x) - \mathbf{\hat{I}}(p_y) ||  + \frac{m}{S} || p_x-p_y ||. 
\end{aligned}  \label{eq:comdist}
\end{equation}

\noindent From~\eqref{eq:comdist}, the parameter $m$ controls the compactness of superpixels whilst $S$ scales the spatial distance and, for a image with $N$ pixels, $S = \sqrt{N/K}$ as in the MSLIC algorithm.

\subsubsection{Spectral Merging} In the original MSLIC algorithm when the area of a seed $s_i$ is below a threshold it is randomly merged with a neighbouring seed $s_j$. However, in our implementation we instead choose the neighbouring seed which satisfies:
\begin{equation}
    j = \operatorname*{arg min}_{s_j \in \mathcal{N}}   || \mathcal{P}^m_i - \mathcal{P}^m_j ||.
\end{equation}
where $\mathcal{P}^m_i$ is the average spectral information of the seed $s_i$ and $\mathcal{N}$ is the set of neighbouring seeds. We choose to merge superpixels, which are most similar in their spectral properties, as this yields a better form of adaptation to the hyperspectral data. 

These proposed changes produce accurate superpixels for HSIs. For further details on our approach, refer to Section II of the supplementary material.

\subsection{Feature Extraction}

Now we seek to extract meaningful features from the extracted superpixels ready for graph construction. In this paper, we use the same features as we did in our previous work on superpixels \cite{OldPaper}. From each superpixel $\mathcal{S}_i$ we extract three different features.
To extract localised spatial information we apply a mean filter to each superpixel to produce a mean feature vector $\vec{\mathcal{S}}^m_i$ which is defined as . 
\begin{equation}
	\vec{\mathcal{S}}^m_i = \frac{ \sum_{j=1}^{n_i} \mathbf{\hat{I}}(p_{i,j}) }{n_i}.
\end{equation}

Using a weighted combination of the mean feature vectors of a superpixel's adjacent neighbours, we can obtain a measure of the spatial information between superpixels. Note that adjacency is defined based on 4-connectivity on the image grid. For each superpixel $\mathcal{S}_i$, we define the set $\mathcal{Z}_i = \{ z_1 , z_2 .. , z_J \}$ which contains the $J$ indexes of its adjacent superpixels. From this, we construct the weighted feature vector $\vec{\mathcal{S}}_i^w$ which reads:
\begin{equation}
		\vec{\mathcal{S}}_i^w = \sum_{j=1}^{J} w_{i,z_j} \vec{\mathcal{S}}_{z_j}^m,  
\end{equation}
where the weight between adjacent superpixels $w_{i,z_j}$ is defined as: 
\begin{equation}
		w_{i,z_j} = \frac{ \exp \left(  -||  \vec{\mathcal{S}}_{z_j}^m - \vec{\mathcal{S}}_{i}^m ||_2^2 / h  \right) }{ \sum_{j=1}^{J} \exp \left(  -||  \vec{\mathcal{S}}_{z_j}^m - \vec{\mathcal{S}}_{i}^m ||_2^2 / h  \right)},
\end{equation}
with $h$ as a predefined scalar parameter. Finally, we propose to extract the centroidal location of each superpixel $\vec{\mathcal{S}}^p_i$ which we calculate as: \begin{equation}
	\vec{\mathcal{S}}^p_i = \frac{ \sum_{j=1}^{n_i} p_{i,j} }{n_i}.
\end{equation}

\subsection{Graph based Classification}
After defining how to get our superpixel set and extracted features, we now turn to explain how we create our weighted graph-representation. However, we first give some background into challenges associated with the computational implementation of graph-based methods and how superpixels can be used to overcome some of these.

As noted by Camps-Valls et al~\cite{CampValls}, many graphical algorithms rely on calculating and manipulating large kernel matrices formed by the labelled and unlabelled data. As an example, for an image with $n$ pixels the associated graph Laplacian is a matrix of size $n \times n$. If we seek to inverse the graph Laplacian via singular value decomposition then the computational complexity would be $O(n^3)$, ruining the scaling that we seek. 

Approximation methods do exist to speed up such matrix inversions. One commonly used technique is the Nystr{\"o}m extension~\cite{NystromOriginal} and it is regularly used to speed up matrix calculations~\cite{CampValls}~\cite{bertozzi}. However, the Nystr{\"o}m extension has several drawbacks. It is unsuitable for sparse applications as the Nystr{\"o}m extension acts as an approximation for complete matrices. 

In this paper, we implement a novel solution to increase the speed and reduce the complexity of graphical classifiers applied to HSIs. Instead of having a graphical representation where each node represents a pixel, we instead use our segmented superpixels as the node set. This greatly reduces the size of our node set as $K \ll n$ and allows us to perform matrix inversion and other calculations without approximations such as the Nystr{\"o}m extension. Furthermore, a superpixel representation should help to boost the classification accuracy as we are defining strong local regions in our data. 
Therefore, from these previously discussed features and our superpixel node set, a weighted, undirected graph $G =(V,E,W)$ can be created. The weight between two connected superpixels $\mathcal{S}_i$ and $\mathcal{S}_j$ is constructed based on two Gaussian kernels  and is given as

\begin{equation}
w_{ij} = s_{ij}l_{ij},
\end{equation}
where 
\begin{equation}
s_{ij} =   \exp\left( \frac{\left( \beta-1 \right)||  \vec{\mathcal{S}}^{w}_i - \vec{\mathcal{S}}^{w}_j ||_2^2  - \beta||\vec{\mathcal{S}}^m_i - \vec{\mathcal{S}}^m_j ||_2^2}{\sigma_s^2}  \right) ,
\end{equation}
\begin{equation}
l_{ij} =   \exp \left( \frac{-||\vec{\mathcal{S}}^{p}_i - \vec{\mathcal{S}}^{p}_j ||_2^2}{\sigma_l^2}\right).
\end{equation}
where $\beta$ balances the influence between the mean and weighted features and $\sigma_s , \sigma_l$ determine the width of the Gaussian kernels. Note that weights are limited in value between $ [ 0,1 ]$ with $1$ implying most similar.  The edge set is constructed using $k$-nearest neighbours. Therefore, the edge weights are defined as:
\begin{equation}
  W_{ij}=\begin{cases}
    w_{ij}, & \text{if $i$ is one of the $k$ nearest neighbor of $j$,}\\
     & \text{or vice versa.}\\
    0 & \text{otherwise}.
  \end{cases}
\end{equation}

In the training stage of the algorithm, a set of labelled spectral pixels are randomly selected from the original HSI. The initial label of each superpixel is taken as the average initial label of its corresponding set of pixels. If no pixel within a superpixel is initially labelled then the superpixel is initially unlabelled.  The labelling information for the superpixels are specified using a matrix $Y \in \mathbb{R}^{K \times c}$, where $c$ is the number of classes present and $K$ is the number of superpixels. $Y_{vl}$ specifies the value of the seed label $l$ for node $v$.  The weight matrix and the initial labelling are then passed into Local and Global Consistency (LCG) algorithm~\cite{LGC}.

LCG is a graph based SSL approach that formalises the smoothness and clustering assumptions of semi-supervised learning by designing a classification function which is smooth upon the graphical structure generated by all the data. The final labelling is specified using a matrix $F \in \mathbb{R}^{K \times c}$. The cost function associated with the matrix $F$ is given by

\begin{equation}
    \mathcal{Q}(F) = \frac{1}{2}   \sum_{i,j =1}^{n} W_{ij} \left|\left| \frac{F_i}{\sqrt{D_{ii}}} - \frac{F_j}{\sqrt{D_{jj}}} \right|\right|^2
    + \frac{\mu}{2} \sum_{i=1}^{n} || F_i - Y_i ||^{2} ,
\end{equation}

where $\mu>0$ is a regularisation parameter. $\mathcal{F}$ denotes the set of $n \times c$ matrices with non-negative entries. The labelling matrix is given by  $F^{*} = \operatorname*{argmin}_{F \in \mathcal{F}} \mathcal{Q}(F)$.

The first term in the cost function is the smoothess constraint, which encourages connected nodes to have similar labelling, whilst the second term fits the finally labelling to  the initially labelled data. Balance between these constraints is set by the parameter $\mu$. The above cost function has a closed form solution which reads: $ F^* = \beta \left(I  - \alpha D^{-\frac{1}{2}}WD^{-\frac{1}{2}} \right)Y$, where 
$\beta = \frac{\mu}{1+\mu}$ and $\alpha = 1 - \beta$.
The final labelling of the nodes is then computed as:  $  y_i = \operatorname*{argmax}_{j \leq c} F_{ij}$. The superpixel labels and the superpixel segmentation are used to construct the final pixel classification map.


\begin{figure}
	\centering
		\includegraphics[width=0.5\textwidth]{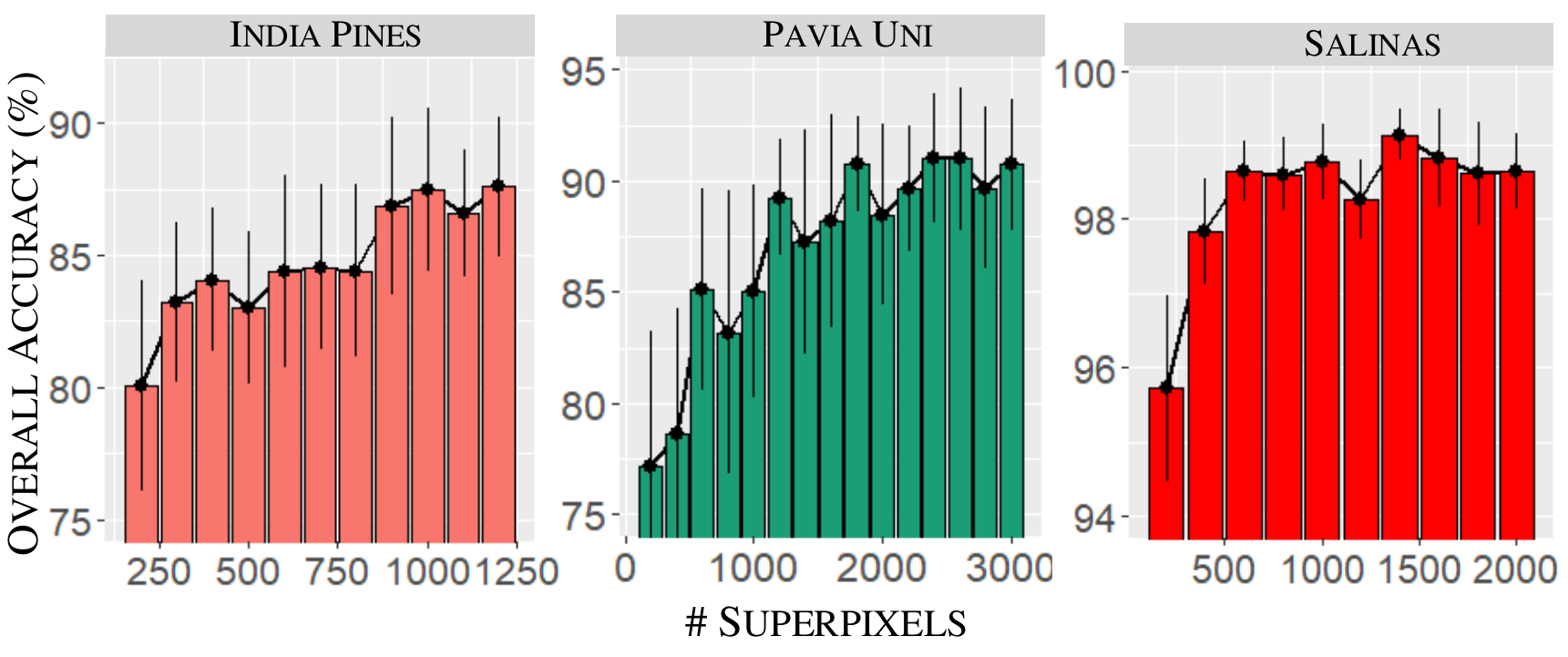} 
	\caption{Sensitivity analysis of the parameter $K$, the number of superpixels, for (a) Indian Pines, (b) Pavia University  nd (c) Salinas. Each data point is the accuracy average of ten repetitions whilst the error bars reflect one standard deviation. For all three data sets the accuracy increases with increasing values of $K$. However, once the number of superpixels is high enough to accurately over-segment the image, there are diminishing returns for increasing the number of superpixels as the accuracy flattens out.}
	\label{fig:sensitivity}
\end{figure}

\section{Experimental Results}
In this section, we detail the experiments conducted to validate the proposed approach.

\subsection{Data Description}
We use three benchmark HSI datasets to evaluate our approach, which have the following characteristics. 

\begin{itemize}
\item  \textbf{Indian Pines Dataset.} The dataset was collected by an airborne visible/infrared imaging spectrometer (AVIRIS) sensor over an agricultural site in Indiana and has 16 classes. The data set consists of $145 \times 145$ pixels, $200$ spectral channels, a spectral range of $0.4$ to $2.5 \mu$m and a spatial resolution of $20$m. 

\item \textbf{Salinas.} This image was also collected by the AVIRIS sensor over Salinas Valley, California, and contains 16 classes. The data set size is $512 \times 217$ pixels and identical to Indian Pines has $200$ spectral channels over $0.4$ to $2.5 \mu$m. The data set is characterised by a high spatial resolution pf $3.7$m per pixel. 

\item \textbf{University of Pavia.} This dataset was acquired by the reflective optics system imaging spectrometer (ROSIS). The image ($610 \times 340$ pixels) covers the Engineering School at the University of Pavia and has 9 classes. The image contains 115 spectral channels from $0.43$ to $0.86 \mu$m and has a has a spatial resolution of $1.3$m.
\end{itemize}

In section III of the supplementary material, we further describe the datasets used and the mathematical background of the evaluation criteria. 

\subsection{Evaluation Protocol}
For all experiments carried out in this paper, each one is repeated $10$ times and the average and standard deviation are provided for each measurement. The number of principal components used were set by demanding that the total explained variance ratio was $ \geq 0.999$. To evaluate the performance of each HSI classifier, we use three commonly implemented evaluation criteria \textit{Overall Accuracy (OA)}, \textit{Average Accuracy (AA)} and the \textit{Kappa Coefficient}. 

To validate the performance of our proposed classification framework SGL, several state-of-the-art HSI classification methods have been implemented to act as comparisons. These are local co-variance matrix representation (LCMR) \cite{LCMR}, superpixel-based classification via multiple kernels (SC-MK) \cite{SCMK}, the edge preserving filter based method (EPF) \cite{EPF}, local binary patterns (LBP) \cite{LBP}, an SVM method \cite{melgani2004classification}  and image fusion and recursive filtering (IFRF) \cite{IFRF}. 

\begin{figure*}[!t]	
	\centering
	\includegraphics[scale=0.45]{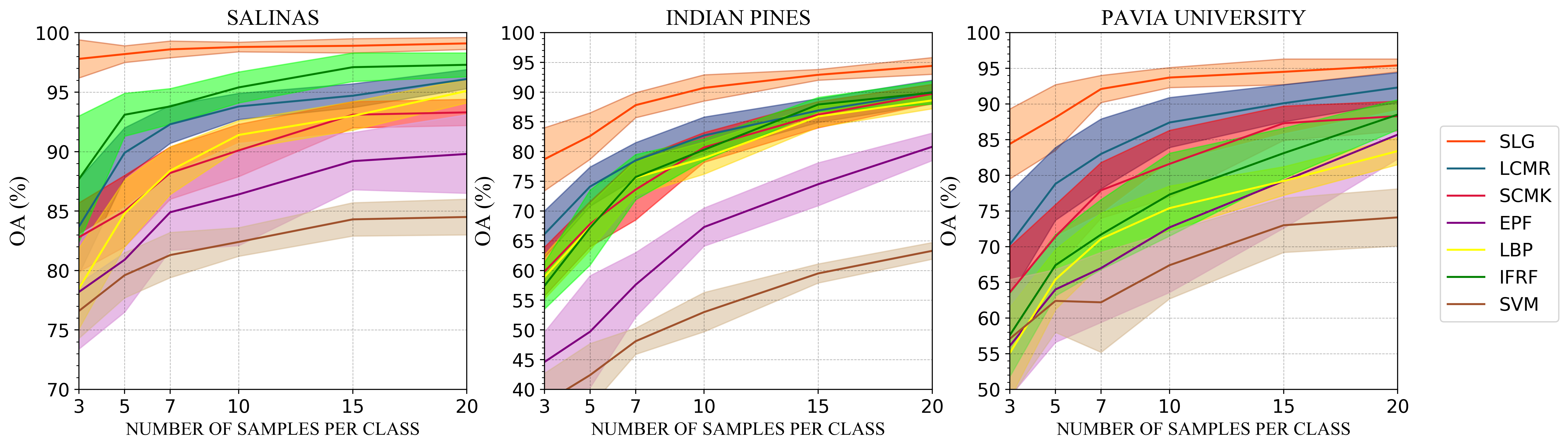}
	\caption{Comparison of the classification accuracy of different methods with varying number of training samples. The methods used are LCMR \cite{LCMR}, SC-MK \cite{SCMK}, EPF \cite{EPF}, LBP \cite{LBP}, IFRF \cite{IFRF}, SVM \cite{melgani2004classification} and the proposed SGL method. The solid lines represent the average of the different methods whilst the shaded area covers one standard deviation from the mean.}
	\label{GraphComparison}
\end{figure*}

\begin{table}[]
\caption{The parameter values used for all experiments in this paper. Note that $\{x,y\}$ signifies a random uniform distribution between $x$ and $y$.}
\begin{tabular}{|cccc|}
\hline
\multicolumn{4}{|c|}{\cellcolor[HTML]{EFEFEF}\textsc{Fixed Parameters}}                                                                                   \\ \hline
\multicolumn{1}{|c}{Parameter} & \multicolumn{2}{c}{Description} & \multicolumn{1}{c|}{Value}  \\ \hline
$m$   & \multicolumn{2}{c}{Controls the compactness of superpixels }    & 10.0    \\
 $h$  & \multicolumn{2}{c}{ Weighted filtering kernel}   & 15.0   \\
 $\sigma_s$ & \multicolumn{2}{c}{Kernel parameter for constructing $s_{ij}$}     &  0.20     \\
 $k$    & \multicolumn{2}{c}{$k$-NN construction}    &    8   \\
  $\mu$   & \multicolumn{2}{c}{ Weighting in the LGC classifier}         &     \{0.1,0.15\}    \\ \hline \hline
\multicolumn{4}{|c|}{\cellcolor[HTML]{EFEFEF}\textsc{Data-based parameters}}                                                                         \\ \hline
\multicolumn{1}{|c}{Parameter} & \multicolumn{1}{c}{Indian Pines} & \multicolumn{1}{c}{Salinas} & \multicolumn{1}{c|}{Pavia Univerisity} \\  \hline
$\beta$     &  0.9         & 0.9     & 0.1             \\ 
    $\sigma_l$  &  \{0.4,0.5\}  & \{3.2.4.0\} &  \{17,20\} \\
    $K$        & $1200$  & $1400$ & $2400$  \\ \hline
\end{tabular}
\label{tab:parameters}

\end{table}

\subsection{Parameter Selection}
In our proposed framework, there are eight hyperparameters that come from the four tasks of our framework.
\begin{itemize}
    \item[] {Superpixel construction:} $K$ and $m$.
    \item[] {Feature Extraction:}  $h$.
    \item[] {Graph construction:} $\sigma_s$, $\sigma_l$, $k$ and $\beta$.
    \item[] {LGC classification:} $\mu$.
\end{itemize}

\begin{table*}[!t]
	\centering
	\caption{OA (\%) of Ten Repeated Experiments with Differing Numbers of training samples per class}
	
	\begin{tabular}{|c|c|c|c|c|c|c|c|}
		\hline
		\multicolumn{8}{|c|}{{\cellcolor[HTML]{EFEFEF}\textsc{Salinas}}}  \\\hline		
		{\cellcolor[HTML]{EFEFEF}\textsc{Samples per Class}}  & {\cellcolor[HTML]{EFEFEF}\textsc{\textbf{{OURS}}}} & {\cellcolor[HTML]{EFEFEF}\textsc{LCMR \cite{LCMR}}} & {\cellcolor[HTML]{EFEFEF}\textsc{SC-MK \cite{SCMK}}}  & {\cellcolor[HTML]{EFEFEF}\textsc{EPF \cite{EPF}}} & {\cellcolor[HTML]{EFEFEF}\textsc{LBP \cite{LBP}}} & {\cellcolor[HTML]{EFEFEF}\textsc{IFRF \cite{IFRF}}} & {\cellcolor[HTML]{EFEFEF}\textsc{SVM \cite{melgani2004classification}}}\\ \hline
		3  & \cellcolor{green}98.0 $\pm$ 0.8\% & 83.7 $\pm$ 4.0\% & 82.8 $\pm$ 2.9\% & 78.2 $\pm$ 4.8\% & 78.5 $\pm$ 3.4\%  & 87.7 $\pm$ 5.3\%  & 76.6 $\pm$ 2.3\% \\ \hline
		5  & \cellcolor{green}99.1 $\pm$ 0.5\% & 89.9 $\pm$ 2.1\% & 85.0 $\pm$ 3.0\% & 80.9 $\pm$ 4.4\% & 84.8 $\pm$ 2.8\% &  93.1 $\pm$ 1.8\% & 79.6 $\pm$ 1.9\%\\ \hline
		7  & \cellcolor{green}99.1 $\pm$ 0.3\% & 92.3 $\pm$ 1.6\% & 88.2 $\pm$ 2.2\% & 84.9 $\pm$ 3.2\% & 88.4 $\pm$ 2.0\%  & 93.8 $\pm$ 1.5\%  & 81.3 $\pm$ 1.9\% \\ \hline
		10 & \cellcolor{green}99.0 $\pm$ 0.4\% & 93.8 $\pm$ 1.1\% & 90.1 $\pm$ 2.2\% & 86.4 $\pm$ 4.3\% & 91.4 $\pm$ 1.2\% & 95.4 $\pm$ 1.3\% & 82.4 $\pm$ 1.2\%\\ \hline
		15 & \cellcolor{green}99.1 $\pm$ 0.3\% & 94.7 $\pm$ 1.0\% & 93.1 $\pm$ 1.1\% & 89.2 $\pm$ 2.4\% & 93.0 $\pm$ 1.2\% & 97.1 $\pm$ 1.2\% & 84.3 $\pm$ 1.4\%\\ \hline
		20 & \cellcolor{green}99.3 $\pm$ 0.2\% & 96.1 $\pm$ 0.8\% & 93.3 $\pm$ 1.1\% & 89.8 $\pm$ 3.3\% & 95.1 $\pm$ 1.1\% & 97.3 $\pm$ 1.0\% & 84.5 $\pm$ 1.5\%  \\ \hline  
		\noalign{\vskip 3mm}
	\end{tabular}
	
	\begin{tabular}{|c|c|c|c|c|c|c|c|}
			\hline
			\multicolumn{8}{|c|}{{\cellcolor[HTML]{EFEFEF}\textsc{Indiana Pines}}}  \\\hline		
			{\cellcolor[HTML]{EFEFEF}\textsc{Samples per Class}} & {\cellcolor[HTML]{EFEFEF}\textsc{\textbf{OURS}}} & {\cellcolor[HTML]{EFEFEF}\textsc{LCMR \cite{LCMR}}} & {\cellcolor[HTML]{EFEFEF}\textsc{SC-MK \cite{SCMK}}}  & {\cellcolor[HTML]{EFEFEF}\textsc{EPF \cite{EPF}}} & {\cellcolor[HTML]{EFEFEF}\textsc{LBP \cite{LBP}}} & {\cellcolor[HTML]{EFEFEF}\textsc{IFRF \cite{IFRF}}} & {\cellcolor[HTML]{EFEFEF}\textsc{SVM \cite{melgani2004classification}}}\\ \hline
			3  & \cellcolor{green}78.7 $\pm$ 5.3\% & 66.1 $\pm$ 3.9\% & 59.8 $\pm$ 4.1\% & 44.6 $\pm$ 5.0\% & 58.9 $\pm$ 3.7\%  & 57.4 $\pm$ 3.9\% & 37.7 $\pm$ 5.0\%\\ \hline
			5  & \cellcolor{green}82.6 $\pm$ 3.9\% & 74.1 $\pm$ 3.3\% & 67.8 $\pm$ 3.8\% & 49.7 $\pm$ 9.4\% & 67.3 $\pm$ 3.9\% &  67.2 $\pm$ 6.3\% & 42.4 $\pm$ 5.3\%\\ \hline
			7  & \cellcolor{green}87.8 $\pm$ 2.1\% & 78.5 $\pm$ 3.0\% & 73.6 $\pm$ 5.1\% & 57.6 $\pm$ 5.4\% & 75.6 $\pm$ 2.9\%  & 75.7 $\pm$ 3.8\%  & 48.1 $\pm$ 2.2\%\\ \hline
			10 & \cellcolor{green}90.7 $\pm$ 2.2\% & 82.7 $\pm$ 3.1\% & 80.7 $\pm$ 2.5\% & 67.3 $\pm$ 3.2\% & 78.9 $\pm$ 2.7\%  & 80.3 $\pm$ 1.8\% & 53.0 $\pm$ 3.3\% \\ \hline
			15 & \cellcolor{green}92.9 $\pm$ 0.9\% & 86.9 $\pm$ 2.0\% & 86.2 $\pm$ 2.2\% & 74.5 $\pm$ 3.6\% & 85.9 $\pm$ 1.8\% &  87.9 $\pm$ 1.2\% & 59.5 $\pm$ 1.6\% \\ \hline
			20 & \cellcolor{green}94.4 $\pm$ 1.4\% & 90.0 $\pm$ 2.0\% & 89.7 $\pm$ 1.6\% & 80.8 $\pm$ 2.3\% & 88.6 $\pm$ 1.4\% & 89.9 $\pm$ 1.9\% & 63.3 $\pm$ 1.4\%\\ \hline	
		    \noalign{\vskip 3mm}
	\end{tabular}

	\begin{tabular}{|c|c|c|c|c|c|c|c|}
			\hline
			\multicolumn{8}{|c|}{{\cellcolor[HTML]{EFEFEF}\textsc{University of Pavia}}}  \\\hline		
			{\cellcolor[HTML]{EFEFEF}\textsc{Samples per Class}} & {\cellcolor[HTML]{EFEFEF}\textsc{\textbf{OURS}}} & {\cellcolor[HTML]{EFEFEF}\textsc{LCMR \cite{LCMR}}} & {\cellcolor[HTML]{EFEFEF}\textsc{SC-MK \cite{SCMK}}}  & {\cellcolor[HTML]{EFEFEF}\textsc{EPF \cite{EPF}}} & {\cellcolor[HTML]{EFEFEF}\textsc{LBP \cite{LBP}}} & {\cellcolor[HTML]{EFEFEF}\textsc{IFRF \cite{IFRF}}} & {\cellcolor[HTML]{EFEFEF}\textsc{SVM \cite{melgani2004classification}}}\\ \hline
			3  & \cellcolor{green}84.4 $\pm$ 4.9\% & 70.3 $\pm$ 7.3\% & 63.6 $\pm$ 6.4\% & 56.1 $\pm$ 7.1\% & 55.1 $\pm$ 6.4\%  & 57.6 $\pm$ 5.8\% & 57.1 $\pm$ 8.3\%\\ \hline
			5  & \cellcolor{green}88.1 $\pm$ 4.6\% & 78.8 $\pm$ 5.1\% & 71.4 $\pm$ 4.5\% & 64.0 $\pm$ 7.4\% & 65.4 $\pm$ 4.2\% &  67.4 $\pm$ 4.3\% & 62.4 $\pm$ 4.4\% \\ \hline
			7  & \cellcolor{green}92.1 $\pm$ 1.9\% & 83.0 $\pm$ 4.9\% & 77.9 $\pm$ 3.9\% & 67.0 $\pm$ 7.6\% & 71.1 $\pm$ 3.6\%  & 71.7 $\pm$ 4.9\% & 62.2 $\pm$ 7.0\%  \\ \hline
			10 & \cellcolor{green}93.7 $\pm$ 1.4\% & 87.4 $\pm$ 3.5\% & 81.6 $\pm$ 4.7\% & 72.7 $\pm$ 9.1\% & 75.4 $\pm$ 3.1\% & 77.3 $\pm$ 5.8\% & 67.4 $\pm$ 4.7\%\\ \hline
			15 & \cellcolor{green}94.5 $\pm$ 1.8\% & 90.1 $\pm$ 2.6\% & 87.3 $\pm$ 2.4\%& 79.2 $\pm$ 6.6\% & 79.2 $\pm$ 2.0\% & 83.1 $\pm$ 3.5\% & 73.0 $\pm$ 3.8\% \\ \hline
			20 & \cellcolor{green}95.4 $\pm$ 0.9\% & 92.3 $\pm$ 2.1\% & 88.3 $\pm$ 2.1\% & 85.7 $\pm$ 3.4\% & 83.4 $\pm$ 1.9\% & 88.5 $\pm$ 2.1\% & 74.1 $\pm$ 4.0\% \\ \hline
	\end{tabular}
	\label{tab::several}
\end{table*}

For the superpixels construction step, we set the ratio of the number of pixels to the number of superpixels $\frac{N}{K}$ must be at least $15$. The parameters $m$, $h$, $\sigma_s$, $k$ and $\mu$ have the same value for all datasets used. These values were found using empirical testing in a coarse to fine search method. The other three parameters, $\sigma_l$, $\beta$ and $K$, change value depending on the HSI used. The parameter values used in the experiments are given in Table~\ref{tab:parameters}.

We leave a discussion of the parameters $\beta$ and $\sigma_l$ to section III of the supplementary material and focus here on the superpixel number $K$. Given that we are using a superpixel based classifier, it is critically important to understand how the superpixel number $K$ effects the accuracy. This is especially true when it is unclear what value of $K$ to pick for a given image. To investigate the effect of changing the parameter $K$, we classified all three HSIs using a varying number of superpixels and $7$ randomly selected samples from each class and plotted the classification accuracy against the superpixel number.  The results for this analysis are given in Fig.~\ref{fig:sensitivity}. In general the classification accuracy increases with the number of superpixels, due to the underlying over-segmentation being more accurate. However, once the image is accurately over-segmented, there are diminishing returns for further increasing the superpixel number. Combined with the fact that increasing the number of superpixels increases the size of the graph and thus the running time, we used the smallest number superpixels that reliably gave a good classification accuracy for each HSI. 

For the compared methods the parameters were set using the default values provided in the demo code or referenced in the papers themselves. The SVM method was implemented using the LIBSVM \cite{LIBSVM} library and uses a Gaussian kernel and five-fold cross validation.

\subsection{Experimental Results}
Our experiments are organised into two parts. Firstly, we compare the classification accuracy of our proposed framework with the comparison classifiers detailed above. Due to the semi-supervised nature of our method, we will be testing the classification performance using very limited amounts of training data. Secondly, we will seek to use visual classification maps to understand and explain the performance of our classifier to relation to the other classifiers.

\begin{table*}[!t]
    \caption{OA (\%) AA (\%) and Kappa (\%) of ten consecutive experiments with ten training samples per class}
    \centering
    \begin{tabular}{|c|c|c|c|c|c|c|c|}
        \hline
        \multicolumn{8}{|c|}{{\cellcolor[HTML]{EFEFEF}\textsc{Salinas}}}  \\\hline	
        {\cellcolor[HTML]{EFEFEF}\textsc{Technique}} &  {\cellcolor[HTML]{EFEFEF}\textsc{\textbf{OURS}}}  & {\cellcolor[HTML]{EFEFEF}\textsc{LCMR \cite{LCMR}}} & {\cellcolor[HTML]{EFEFEF}\textsc{SC-MK \cite{SCMK}}} & {\cellcolor[HTML]{EFEFEF}\textsc{EPF \cite{EPF}}} & {\cellcolor[HTML]{EFEFEF}\textsc{LBP \cite{LBP}}} & {\cellcolor[HTML]{EFEFEF}\textsc{IFRF \cite{IFRF}}} & {\cellcolor[HTML]{EFEFEF}\textsc{SVM \cite{melgani2004classification}}}\\ \hline
        OA & \cellcolor{green}99.24 $\pm$ 0.16\% & 93.90 $\pm$ 1.29\% & 90.38 $\pm$ 2.42\% & 86.53 $\pm$ 1.99\% & 90.68 $\pm$ 1.35\% & 95.87 $\pm$ 1.62\% & 82.42 $\pm$ 1.15\% \\
        AA & \cellcolor{green}98.90 $\pm$ 1.51\% & 96.26 $\pm$ 1.02\% & 94.16 $\pm$ 1.11\% & 93.51 $\pm$ 0.91\% & 93.00 $\pm$ 1.03\% & 96.24 $\pm$ 1.43\% & 88.55 $\pm$ 0.99\% \\
        Kappa & \cellcolor{green}99.15 $\pm$ 0.17\% & 93.22 $\pm$ 1.44\% & 89.33 $\pm$ 2.67\% & 85.10 $\pm$ 2.15\% & 90.68 $\pm$ 1.36\% & 95.41 $\pm$ 1.80\% & 80.53 $\pm$ 1.25\%\\ \hline 
    \end{tabular}
    
    \begin{tabular}{|c|c|c|c|c|c|c|c|} 
        \hline
        \multicolumn{8}{|c|}{{\cellcolor[HTML]{EFEFEF}\textsc{Indiana Pines}}}  \\\hline	
         {\cellcolor[HTML]{EFEFEF}\textsc{Technique}} & {\cellcolor[HTML]{EFEFEF}\textsc{\textbf{OURS}}}  & {\cellcolor[HTML]{EFEFEF}\textsc{LCMR \cite{LCMR}}}  & {\cellcolor[HTML]{EFEFEF}\textsc{SC-MK \cite{SCMK}}} &  {\cellcolor[HTML]{EFEFEF}\textsc{EPF \cite{EPF}}}  & {\cellcolor[HTML]{EFEFEF}\textsc{LBP \cite{LBP}}}  & {\cellcolor[HTML]{EFEFEF}\textsc{IFRF \cite{IFRF}}} & {\cellcolor[HTML]{EFEFEF}\textsc{SVM \cite{melgani2004classification}}}\\ \hline
        OA & \cellcolor{green}90.89 $\pm$ 2.98\% & 82.74 $\pm$ 2.32\% & 79.91 $\pm$ 2.60\% & 68.95 $\pm$ 2.01\% & 80.52 $\pm$ 2.03\% & 80.86 $\pm$ 3.76\% & 51.20 $\pm$ 3.92\% \\
        AA & \cellcolor{green}92.16 $\pm$ 6.77\% & 90.48 $\pm$ 1.56\% & 87.86 $\pm$ 1.53\% & 71.39 $\pm$ 3.49\% & 88.46 $\pm$ 1.29\% & 74.99 $\pm$ 3.16\% & 51.19 $\pm$ 3.22\%\\
        Kappa & \cellcolor{green}87.50 $\pm$ 3.33\% & 80.51 $\pm$ 2.59\% & 77.31 $\pm$ 2.93\% & 65.02 $\pm$ 2.25\% & 78.09 $\pm$ 2.23\% & 78.45 $\pm$ 4.16\% & 45.41 $\pm$ 4.11\% \\ \hline
    \end{tabular}
    
    \begin{tabular}{|c|c|c|c|c|c|c|c|} 
        \hline
        \multicolumn{8}{|c|}{{\cellcolor[HTML]{EFEFEF}\textsc{University of Pavia}}} \\ \hline	
         {\cellcolor[HTML]{EFEFEF}\textsc{Technique}} &  {\cellcolor[HTML]{EFEFEF}\textsc{\textbf{OURS}}}   & {\cellcolor[HTML]{EFEFEF}\textsc{LCMR \cite{LCMR}}}  & {\cellcolor[HTML]{EFEFEF}\textsc{SC-MK \cite{SCMK}}} &  {\cellcolor[HTML]{EFEFEF}\textsc{EPF \cite{EPF}}}  & {\cellcolor[HTML]{EFEFEF}\textsc{LBP \cite{LBP}}}  & {\cellcolor[HTML]{EFEFEF}\textsc{IFRF \cite{IFRF}}} & {\cellcolor[HTML]{EFEFEF}\textsc{SVM \cite{melgani2004classification}}} \\ \hline
         OA & \cellcolor{green}93.70 $\pm$ 1.35\% &88.29 $\pm$ 4.01\% & 80.23 $\pm$ 4.06\% & 73.92 $\pm$ 7.06\% & 72.66 $\pm$ 4.29\% & 76.36 $\pm$ 3.81\% & 67.40 $\pm$ 4.66\% \\
         AA    & \cellcolor{green}93.25 $\pm$ 5.03\% & 90.72 $\pm$ 1.67\% & 83.99 $\pm$ 2.15\% & 76.10 $\pm$ 5.06\% & 75.99 $\pm$ 2.71\% & 70.39 $\pm$ 3.24\% & 70.08 $\pm$ 2.48\% \\
         Kappa & \cellcolor{green}91.71 $\pm$ 1.73\% & 84.91 $\pm$ 4.89\% & 74.63 $\pm$ 4.60\% & 67.40 $\pm$ 8.23\% & 72.66 $\pm$ 4.25\% & 69.70 $\pm$ 4.55\% & 59.38 $\pm$ 4.88\%\\ \hline
    \end{tabular}

    \label{tab::10}
\end{table*}

\begin{figure*}[t!]
	\centering
	\begin{tabular}{ccccccc}
		\includegraphics[width=30mm,height=48mm]{Salinas} &  
		\includegraphics[width=30mm,height=48mm]{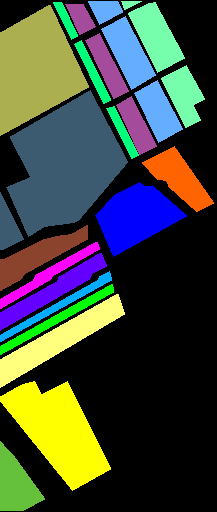} & 
		\includegraphics[width=30mm,height=48mm]{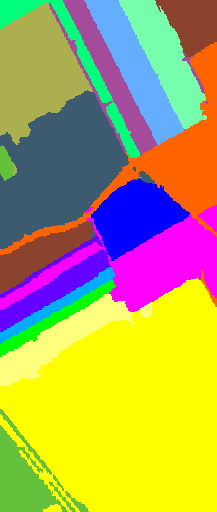} &
		\includegraphics[width=30mm,height=48mm]{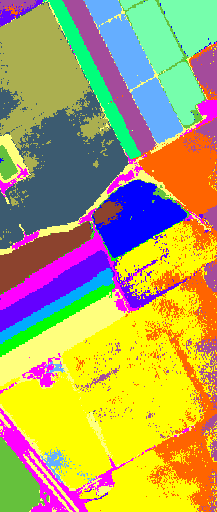} &
		\includegraphics[width=30mm,height=48mm]{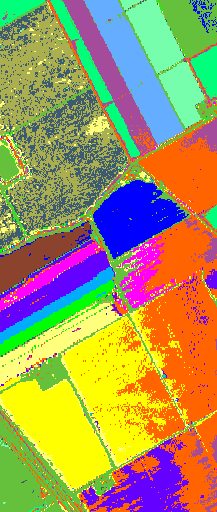} \\
		(a)  & (b) & (c) & (d) & (e) \\[6pt]
		\includegraphics[width=30mm,height=48mm]{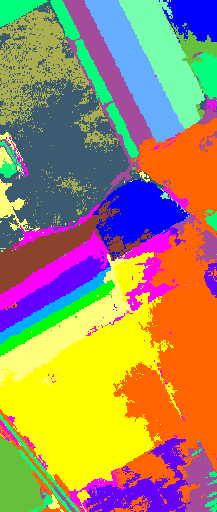} &
		\includegraphics[width=30mm,height=48mm]{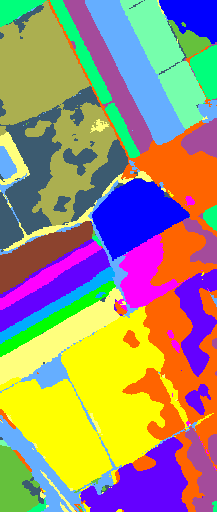} &
		\includegraphics[width=30mm,height=48mm]{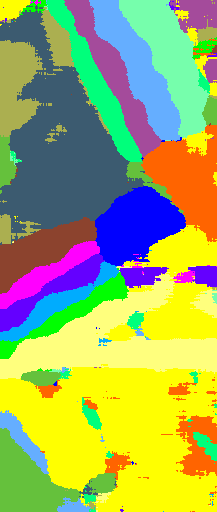} &
		\includegraphics[width=30mm,height=48mm]{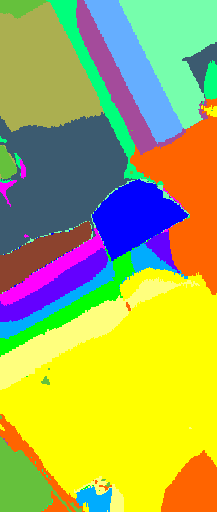} \\
		(f) & (g)& (h) & (i) \\[6pt]	
	\end{tabular}
	\caption{Salinas data set. (a) Colour composite image. (b) Ground truth. (c)-(h) are classifications maps produced using $10$ labelled samples for each class. The methods used were: (c) the proposed SGL , (d) LCMR \cite{LCMR}, (e) SVM \cite{melgani2004classification} , (f) SC-MK\cite{SCMK}, (g) EPF \cite{EPF}, (h) LBP \cite{LBP} and (i) IFRF \cite{IFRF} }
	\label{SC}
\end{figure*}

(E1) In our first experiment, we evaluate the overall accuracy (OA) of our method against the state-of-the-art when using a reduced amount of labelled data for training ($\{3,5,7,10,15,20\}$ randomly selected samples per class). The accuracy of the different classifiers for the three benchmark datasets are given in Table \ref{tab::several} and the graphical representation of the results is shown in Fig. \ref{GraphComparison}.

We see that the accuracy produced by the SGL framework is, by a significant margin, the best of any classifier considered in this paper. The SGL framework produces the best accuracy for all three benchmark images for each differing amount of labelled data. In particular, the average difference in OA between SGL and its nearest competitor LCMR \cite{LCMR}, across the three datasets, was $9\%$  when using $5$ samples per class and was $13.7\%$ when using $3$ samples per class. Highlighting the semi-supervised nature of the SGL framework that allows it to exploit information present in the unlabelled data to overcome the limited amount of labelled samples.

\begin{figure*}[t!]
	\centering
	\begin{tabular}{ccccccc}
		\includegraphics[width=30mm,height=62mm]{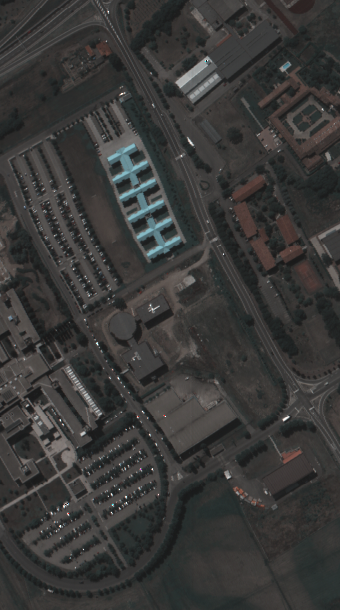} &  
		\includegraphics[width=30mm,height=62mm]{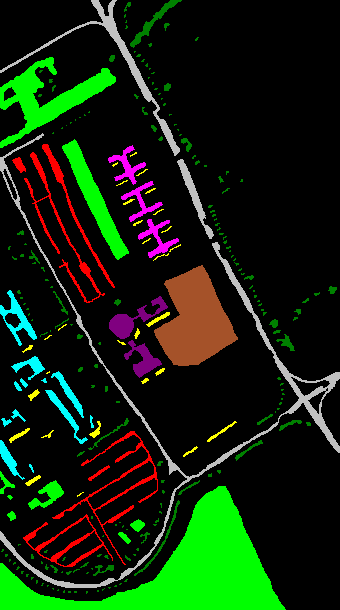} & 
		\includegraphics[width=30mm,height=62mm]{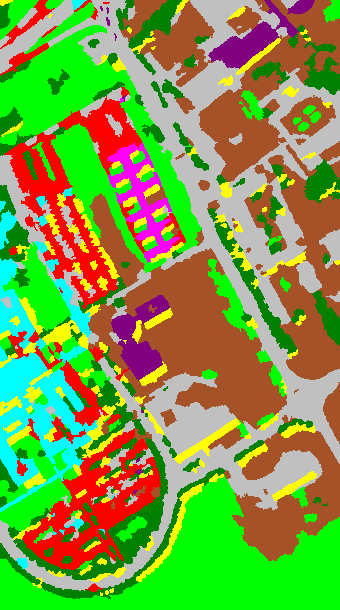} &
		\includegraphics[width=30mm,height=62mm]{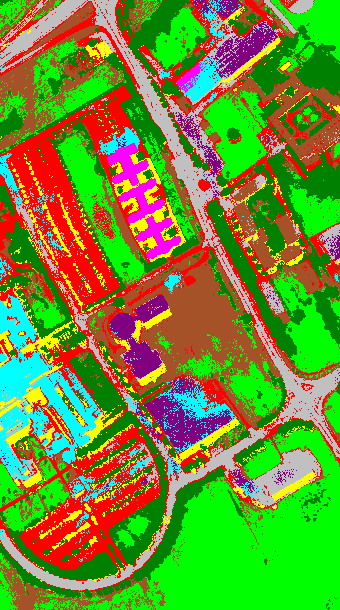} &
		\includegraphics[width=30mm,height=62mm]{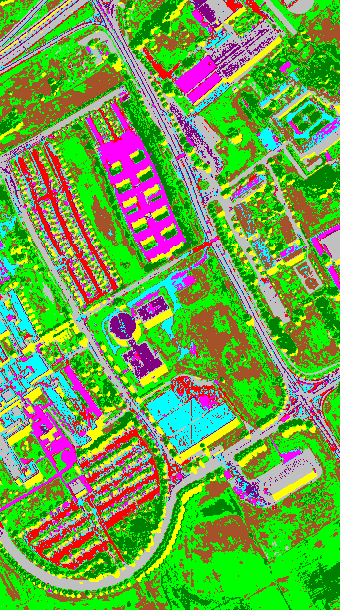} \\
		(a)  & (b) & (c) & (d) & (e)\\[6pt]
		\includegraphics[width=30mm,height=62mm]{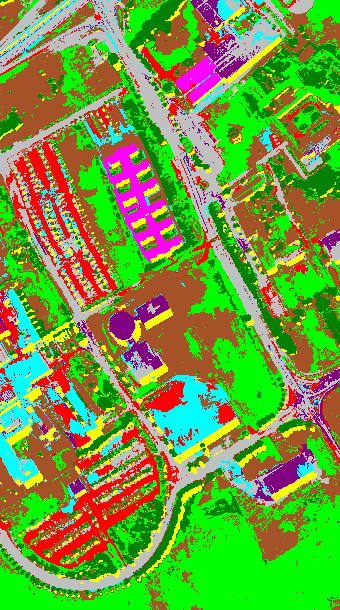} &
		\includegraphics[width=30mm,height=62mm]{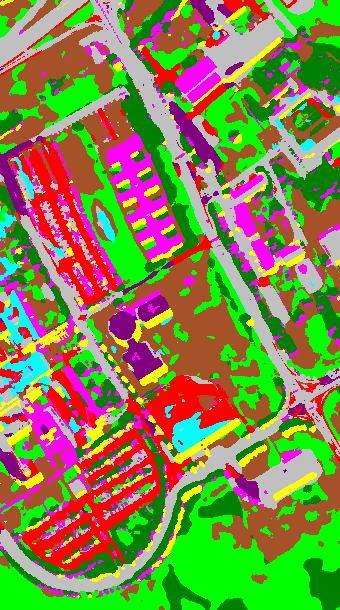} &
		\includegraphics[width=30mm,height=62mm]{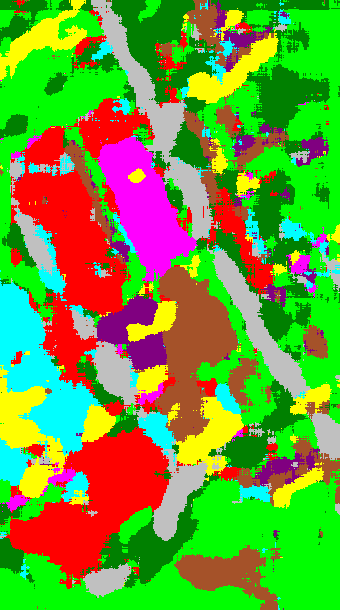} &
		\includegraphics[width=30mm,height=62mm]{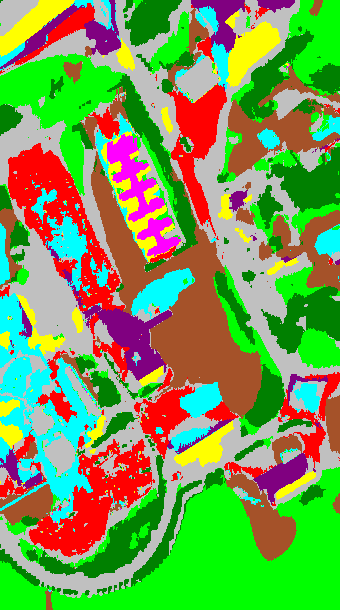} \\
		(f) & (g) & (h) & (i) & \\[6pt]		
	\end{tabular}
	\caption{Pavia University data set. (a) Colour composite image. (b) Ground truth. (c)-(h) are classifications maps produced using $10$ labelled samples for each class. The methods used were: (c) the proposed SGL , (d) LCMR \cite{LCMR}, (e) SVM \cite{melgani2004classification} , (f) SC-MK\cite{SCMK}, (g) EPF \cite{EPF}, (h) LBP \cite{LBP} and (i) IFRF \cite{IFRF} }
	\label{UPC}
\end{figure*}

\begin{figure*}[t!]
	\centering
	\begin{tabular}{ccccc}
		\includegraphics[width=30mm,height=30mm]{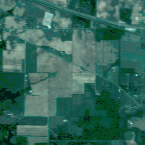} &  
		\includegraphics[width=30mm,height=30mm]{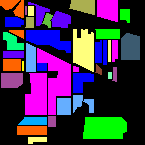} & 
		\includegraphics[width=30mm,height=30mm]{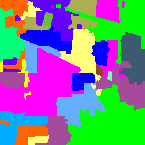} &
		\includegraphics[width=30mm,height=30mm]{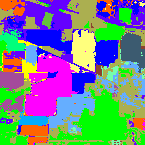} &
		\includegraphics[width=30mm,height=30mm]{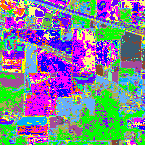} \\
		(a) & (b) & (c) & (d) & (e) \\
		\includegraphics[width=30mm,height=30mm]{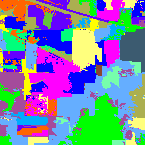} &
		\includegraphics[width=30mm,height=30mm]{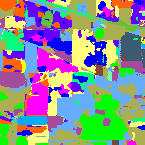} &
		\includegraphics[width=30mm,height=30mm]{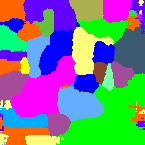} &
		\includegraphics[width=30mm,height=30mm]{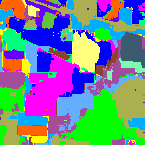} \\
		(f) & (g) & (h) & (i)\\
			
	\end{tabular}
	\caption{Indian pines data set. (a) Colour composite image. (b) Ground truth. (c)-(h) are classifications maps produced using $10$ labelled samples for each class. The methods used were: (c) the proposed SGL , (d) LCMR \cite{LCMR}, (e) SVM \cite{melgani2004classification} , (f) SC-MK\cite{SCMK}, (g) EPF \cite{EPF}, (h) LBP \cite{LBP} and (i) IFRF \cite{IFRF} }
	\label{IPC}
\end{figure*}

(E2) To gain an understanding about how each classifier was performing and the explanation for the large increase in classification accuracy obtained by SGL, we produce visual classification maps. For each HSI we use ten labelled samples per class and calculated the overall accuracy (OA), average accuracy (AA), the Kappa coefficient and the full classification map. The results for this experiment are reported in Table \ref{tab::10}. Furthermore, Fig. \ref{SC}-\ref{IPC} give a colour composite image, ground truth image and the final classification maps for the seven considered methods. In section III of the supplementary material we provide a class by class accuracy breakdown. 

Examining the OA, AA and Kappa coefficient of the differing methods, we observe that SGL is again the best performing method with an average improvement of OA $+12.3\%$, AA $+9.5\%$ and Kappa $+11.6\%$ in the Indian pines scene, OA $+15.4\%$, AA $+13.8\%$ and Kappa $+17.3\%$ in the Pavia University scene and OA $+7.8\%$, AA $+4.3\%$ and Kappa $+8.4\%$ in the Salinas scene compared to the other classifiers (excluding the SVM).

To provide an explanation for the fantastic performance of SGL compared to the other methods let us examine the classification maps. The poorest performing classifier was the SVM. The SVM method only uses spectral information and as a result produces very noisy classification maps. The EPF method seeks to optimise the SVM classification map with an edge preserving filter to smooth out some of this noise and from these results we can see it successfully does so. However, the poor performance of the underlying SVM classification prevents the EPF method from achieving good classification. The LBP and IFRF methods produce over-smooth classification results when only a limited amount of data is available. This causes poor performance in the more complicated Indian Pines and Pavia University images. The LCMR and SCMK methods are the closest competitors to the SGL method with LCMR slightly outperforming the SCMK method due to a slightly higher amount of smoothing. Both of these methods manage to preserve edges and boundaries whilst producing smooth classification maps. This is due to the inclusion of spatial information via local neighbouring pixel construction and superpixel based kernels respectively.  

What sets SGL apart from the other methods considered is that the classification map has been intelligently smoothed with near complete preservation of edges and boundaries. Primarily, this is due to the use of superpixels as the node set in our graph. The superpixels produce by our novel superpixel algorithm have accurately preserved the edges and boundaries in the image. Therefore, when we assign labels to each superpixel, rather than each pixel, we smooth our classification map across the homogeneous superpixels whilst retaining boundaries.  

\section{Conclusion}
In this paper, we have developed a novel semi-semi-supervised graph-based approach, SGL, for the classification of hyperspectral images. The proposed method can be split into three main stages: over-segmentation of HSIs with a novel superpixel algorithm specially designed for dealing with hyperspectral data, feature extraction to extract discriminative features and graph construction and classification. Our experiments with real benchmark HSIs demonstrate that our proposed method greatly outperforms other state-of-the-art classifiers in terms of qualitative and quantitative results, especially when using an incredibly small amount of data.

The semi-supervised nature of our solution exploits data present in the unlabelled data and can overcome the issue of having a highly limited training set, a common problem in the field of remote sensing. Furthermore, for the first time we propose using superpixels as the nodes of a pure graphical classifier which has two large benefits. Firstly, the size of the superpixel graph is much smaller than a pixel based graph allowing for computational reasonable run times without the need for matrix approximations. Secondly, applying labels to superpixels intelligent smooths our classification maps with near perfect preservation of edges and boundaries. 

In our future work , we intend on applying deep learning to automate the extraction of deep features. Furthermore, we seek to apply recent work on heterogeneous graphs to investigate a combined superpixel/pixel representation. 

\section{Acknowledgment}
The authors would like to thank Prof. D. Landgrebe from Purdue University and the NASA Jet Propulsion Laboratory for providing the hyperspectral data sets. We would also like to thank Prof David Coomes from the Department of Plant Sciences, University of Cambridge, for his advice and support. This work was supported by the UK Engineering and Physical Sciences Research Council (EPSRC) and the National Physical Laboratory (NPL). Support from the Centre for Mathematical Imaging in Healthcare (CMIH) University of Cambridge and the Maths in Healthcare Centre is greatly acknowledged. CBS acknowledges support from the Leverhulme Trust project on Breaking the non-convexity barrier, the Philip Leverhulme Prize, the EPSRC grant Nr. EP/M00483X/1, the EPSRC Centre Nr. EP/N014588/1, the European Union Horizon 2020 research and innovation programmes under the Marie Skodowska-Curie grant agreement No 777826 NoMADS and No 691070 CHiPS, the Cantab Capital Institute for the Mathematics of Information and the Alan Turing Institute. We gratefully acknowledge
the support of NVIDIA Corporation with the donation of a Quadro P6000 GPU used for this research.

\bibliographystyle{IEEEtran}
\bibliography{bibliographyV2.bib}

\clearpage
\setcounter{section}{0}
\title{{\LARGE{\textsc{Supplemental Material For the Article:}}} 
\protect\\ \Huge{Superpixel Contracted Graph-Based Learning for Hyperspectral 
Image Classification}}

%
%

\markboth{Journal of \LaTeX\ Class Files,~Vol.~13, No.~9, September~2014}%
{Shell \MakeLowercase{\textit{et al.}}: Bare Demo of IEEEtran.cls for Journals}
%


\maketitle


%
\IEEEpeerreviewmaketitle

\begin{table*}[h]
    \label{tab::class}
	\centering
	\caption{Class breakdown of the three test data sets}
	  \resizebox{\textwidth}{!}{ 
		\begin{tabular}{|c|c|c|c|c|c|c|c|c|c|c|c|}  \hline
			\multicolumn{4}{|c|}{{\cellcolor[HTML]{EFEFEF}\textsc{Indian Pines}}} & \multicolumn{4}{|c|}{{\cellcolor[HTML]{EFEFEF}\textsc{University of Pavia}}} & \multicolumn{4}{|c|}{{\cellcolor[HTML]{EFEFEF}\textsc{Salinas}}}\\ \hline
			{\cellcolor[HTML]{EFEFEF}\textsc{Class}} & {\cellcolor[HTML]{EFEFEF}\textsc{Colour}} & {\cellcolor[HTML]{EFEFEF}\textsc{Name}} & {\cellcolor[HTML]{EFEFEF}\textsc{Number}} & {\cellcolor[HTML]{EFEFEF}\textsc{Class}} & {\cellcolor[HTML]{EFEFEF}\textsc{Colour}} & {\cellcolor[HTML]{EFEFEF}\textsc{Name}} & {\cellcolor[HTML]{EFEFEF}\textsc{Number}} & {\cellcolor[HTML]{EFEFEF}\textsc{Class}} & {\cellcolor[HTML]{EFEFEF}\textsc{Colour}} & {\cellcolor[HTML]{EFEFEF}\textsc{Name}} & {\cellcolor[HTML]{EFEFEF}\textsc{Number}} \\ \hline
			1 & \crule[A1]{0.5cm}{0.2cm} & \textsc{Alfalfa} & 41 & 1 & \crule[B1]{0.5cm}{0.2cm} & \textsc{Asphalt} & 6621 & 1 & \crule[A1]{0.5cm}{0.2cm} & \textsc{Broccoli green weeds 1} & 2004 \\
			2 & \crule[A2]{0.5cm}{0.2cm} & \textsc{Corn-notill} & 1423 & 2 & \crule[B2]{0.5cm}{0.2cm} & \textsc{Meadows} & 18639 & 2 & \crule[A2]{0.5cm}{0.2cm} & \textsc{Broccoli green weeds} 1 & 3721 \\
			3 & \crule[A3]{0.5cm}{0.2cm} & \textsc{Corn-mintill} & 825 & 3 & \crule[B3]{0.5cm}{0.2cm} & \textsc{Gravel} & 2089 & 3 & \crule[A3]{0.5cm}{0.2cm} & \textsc{Fallow} & 1971 \\
			4 & \crule[A4]{0.5cm}{0.2cm} & \textsc{Corn} & 232 & 4 & \crule[B4]{0.5cm}{0.2cm} & \textsc{Trees} & 3054 & 4 & \crule[A4]{0.5cm}{0.2cm} & \textsc{Fallow rough plow} & 1386 \\
			5 & \crule[A5]{0.5cm}{0.2cm} & \textsc{Grass-pasture} & 478 & 5 & \crule[B5]{0.5cm}{0.2cm} & \textsc{Metal sheet} & 1335 & 5 & \crule[A5]{0.5cm}{0.2cm} & \textsc{Fallow smooth} & 2673 \\
			6 & \crule[A6]{0.5cm}{0.2cm} & \textsc{Grass-trees} & 725 & 6 & \crule[B6]{0.5cm}{0.2cm} & \textsc{Bare soil} & 5019 & 6 & \crule[A6]{0.5cm}{0.2cm} & \textsc{Stubble} & 3954 \\
			7 & \crule[A7]{0.5cm}{0.2cm} & \textsc{Grass-pasture-mowed} & 24 & 7 & \crule[B7]{0.5cm}{0.2cm} & \textsc{Bitumen} & 1320 & 7 & \crule[A7]{0.5cm}{0.2cm} & \textsc{Celery} & 3574 \\
			8 & \crule[A8]{0.5cm}{0.2cm} & \textsc{Hay-windrowed} & 473 & 8 & \crule[B8]{0.5cm}{0.2cm} & \textsc{Bricks} & 3672 & 8 & \crule[A8]{0.5cm}{0.2cm} & \textsc{Grapes untrained} & 11266 \\
			9 & \crule[A9]{0.5cm}{0.2cm} & \textsc{Oats} & 15 & 9 & \crule[B9]{0.5cm}{0.2cm} & \textsc{Shadows} & 937 & 9 & \crule[A9]{0.5cm}{0.2cm} & \textsc{Soil vinyard develop} & 6198 \\
			10 & \crule[A10]{0.5cm}{0.2cm} & \textsc{Soybean-notill} &  967 & & & & & 10 & \crule[A10]{0.5cm}{0.2cm} & \textsc{Corn senesced green weeds} & 3273 \\
			11 & \crule[A11]{0.5cm}{0.2cm} & \textsc{Soybean-mintill} & 2450 & & & & & 11 & \crule[A11]{0.5cm}{0.2cm} & \textsc{Lettuce romaine 4wk} & 1063 \\
			12 & \crule[A12]{0.5cm}{0.2cm} & \textsc{Soybean-clean} & 588 & & & & & 12 & \crule[A12]{0.5cm}{0.2cm} & \textsc{Lettuce romaine 5wk} & 1922 \\
			13 & \crule[A13]{0.5cm}{0.2cm} & \textsc{Wheat} & 200 & & & & & 13 & \crule[A13]{0.5cm}{0.2cm} & \textsc{Lettuce romaine 6wk} & 911 \\
			14 & \crule[A14]{0.5cm}{0.2cm} & \textsc{Woods} & 1260 & & & & & 14 & \crule[A14]{0.5cm}{0.2cm} & \textsc{Lettuce romaine 7wk} & 1065 \\
			15 & \crule[A15]{0.5cm}{0.2cm} & \textsc{Buildings-Grass-Trees-Drives} & 381 & & & & & 15 & \crule[A15]{0.5cm}{0.2cm} & \textsc{Vinyard untrained} & 7263 \\
			16 & \crule[A16]{0.5cm}{0.2cm} & \textsc{Stone-Steel-Towers} & 88 & & & & & 16 & \crule[A16]{0.5cm}{0.2cm} & \textsc{Vinyard vertical trellis} & 1802 \\ \hline
		\end{tabular}
		}
	\label{tableset}
\end{table*}

\section{Outline}
\IEEEPARstart{T}he purpose of this supplementary material is to provide further details of the methodology used in the main paper as well as provide additional experimental results to validate the performance of our proposed framework. The supplementary material is divided into two sections:

\begin{itemize}
    \item \textbf{Section \ref{sec::superpixel}.} 
    In this section we give more detail into the hyperspectral extension of \textit{MSLIC} \cite{MSLIC}, named \textit{Hyper-Manifold SLIC} (HSM), which allows it to accurately over-segment hyperspectral images. 
    \item \textbf{Section \ref{sec::further_results}.} 
    In this section we describe the benchmarking datasets in additional detail, further expand the parameter analysis, and provide additional classification results of the SGL method.
\end{itemize}

\section{Superpixel Construction}
\label{sec::superpixel}

As our starting point\textcolor{blue}{,} we used the \textit{Manifold SLIC} method \cite{MSLIC} algorithm developed by Liu et al and we would refer readers to their paper for a detailed explanation of the manifold extension. In the main paper\textcolor{blue}{,} we list the major changes we made to the MSLIC algorithm to produce our hyperspectral extension which we call \textit{Hyperspectral Manifold SLIC} (HMS).  In this section, we list some other additional changes that we made to MSLIC that were not stated in the main paper. We also include additional visual examples of the application of HMS to real HSIs. 

\subsection{Parameter Changes}
From the original paper we made a small number of parameter changes. In this section, we give the reasoning behind these changes. 

\bigskip
\textbf{Convergence Conditions.}  In our implementation we stop the iteration loop when either the residual energy increases or when the percentage decrease in the residual energy is less than $10\%$. This is to prevent the superpixel algorithm for running for extended periods of time. 

\smallskip
\textbf{Enforced Connectivity.} The final superpixels generated by the algorithm must be 4-connected on the image grid. This is ensured by an enforced connectivity algorithm which has two thresholds for the minimum and maximum superpixel size. In our algorithm\textcolor{blue}{,} we use a minimum superpixel size of 8 and a maximum superpixel size of $\frac{10 n}{K}$\textcolor{blue}{,} where $n$ is the number of pixels and $K$ is the initial number of superpixels.  The reason for choosing such a small minimum cluster size was that certain classes in the Indian Pines dataset were tiny in size and we needed to be able to capture these small areas.

\subsection{Supplementary Visual Results for HSM}

\begin{figure*}
	\centering
	\begin{tabular}{ccccc}
		\includegraphics[width=30mm,height=30mm]{Indian_pines} &  
		\includegraphics[width=30mm,height=30mm]{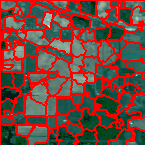} & 
		\includegraphics[width=30mm,height=30mm]{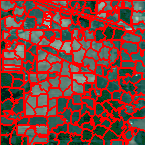}&
		\includegraphics[width=30mm,height=30mm]{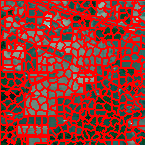} &
		\includegraphics[width=30mm,height=30mm]{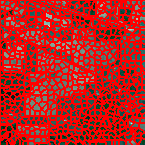} \\
		(a) & (b) & (c) & (d) & (e) \\
	\end{tabular}
	\caption{Superpixel over-segmentations on the Indian Pines scene generated by the HMS extension. From left to right: (a) the composite RGB image, (b)-(e) superpixel segmentations with $129$, $287$, $434$ and $791$ superpixels respectively.}
	\label{IHSM}
\end{figure*}

\begin{figure*}
	\centering
	\begin{tabular}{ccccc}
		\includegraphics[width=32mm,height=68mm]{PaviaU} &  
		\includegraphics[width=32mm,height=68mm]{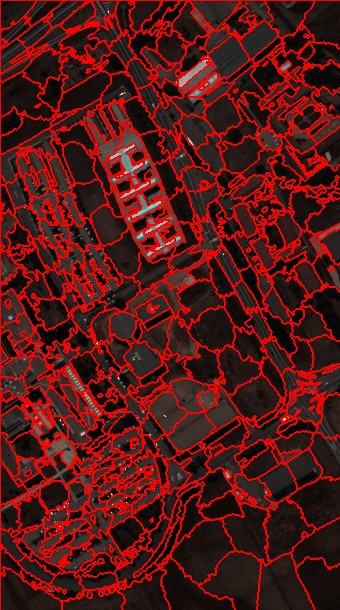} & 
		\includegraphics[width=32mm,height=68mm]{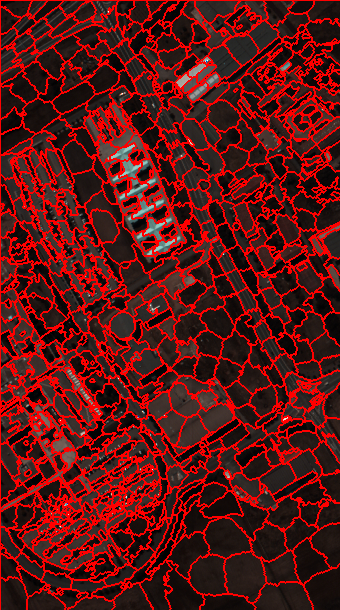} &
		\includegraphics[width=32mm,height=68mm]{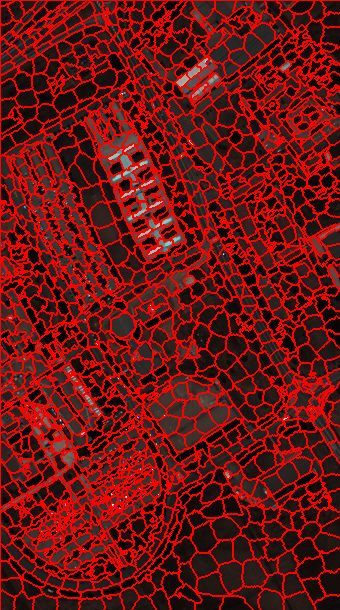} &
		\includegraphics[width=32mm,height=68mm]{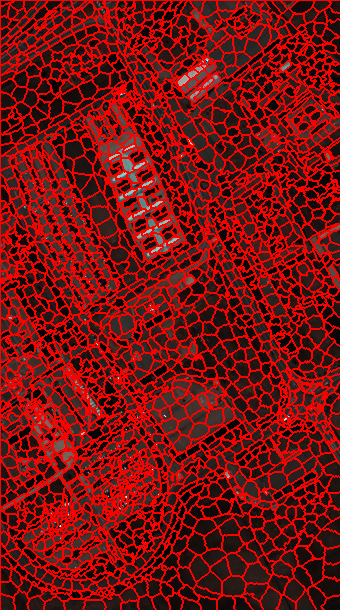} \\
		(a)  & (b) & (c) & (d) & (e)\\[6pt]
	\end{tabular}
	\caption{Superpixel over-segmentations on the University of Pavia scene generated by the HMS extension. From left to right: (a) the composite RGB image, (b)-(e) superpixel segmentations with $948$, $1286$, $1662$ and $1963$ superpixels respectively.}
	\label{UHSM}
\end{figure*}

\begin{figure*}
	\centering
	\begin{tabular}{ccccc}
		\includegraphics[width=32mm,height=58mm]{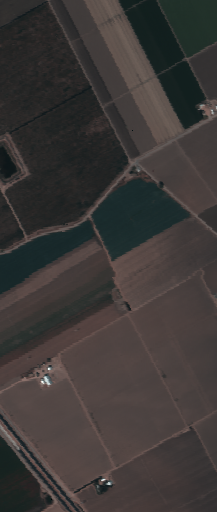} &  
		\includegraphics[width=32mm,height=58mm]{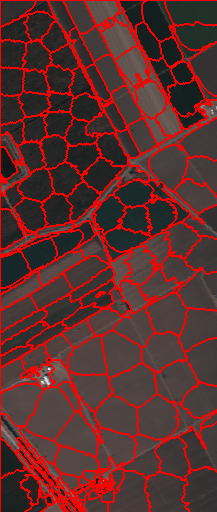} & 
		\includegraphics[width=32mm,height=58mm]{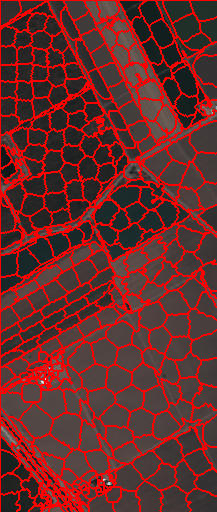} &		
		\includegraphics[width=32mm,height=58mm]{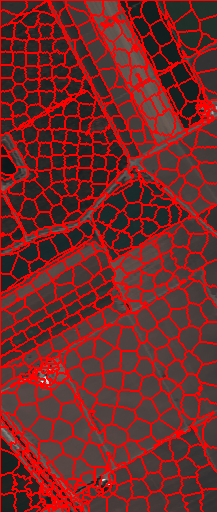} &
		\includegraphics[width=32mm,height=58mm]{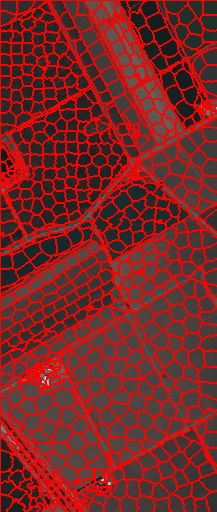} \\
		(a)  & (b) & (c) & (d) & (e)\\[6pt]
	\end{tabular}
	\caption{Superpixel over-segmentations on the Salinas scene generated by the HMS extension. From left to right: (a) the composite RGB image, (b)-(e) superpixel segmentations with $244$, $465$, $639$ and $919$ superpixels respectively.}
	\label{SHSM}
\end{figure*}

In Figs \ref{IHSM}-\ref{SHSM} we provide more examples of the application of HSM to the three different HSIs considered in the main paper. We provide over-segmentations with differing numbers of superpixels highlighting the content sensitivity of the algorithm. In particular\textcolor{blue}{,} note that Indian Pines and Salinas are easily over-segmented using a small number of superpixels whilst the more complex structure of Pavia University requires more superpixels to achieve an accurate over-segmentation.

\section{Supplementary Results}
\label{sec::further_results}

In this section, we expand the details regarding the experimental methodology used. Additionally, we provide further experimental results that validate the performance of our proposed framework SGL.

\subsection{Further Description of the Data Sets}
The three labelled datasets used in the main paper are "AVIRIS Indian Pines", "AVIRIS Salinas" and "Reflective Optics System Imaging Spectrometer (ROSIS-03) University of Pavia." Whilst the main paper describes the format of the three datasets, In this section, we give further details on the data sets and prepossessing. A class by class breakdown of each data set listing the different classes and the number of samples is given in Table \ref{tableset}. 
\medskip

\textbf{Indian Pines} consists of mainly agricultural classes with a small amount of non-organic land cover. Due to presence  The different classes vary greatly in size with the smallest classes in the tens of pixels whilst the largest classes have several thousand pixels. Bands [104-108], [150-163] and 220 were removed prior to classification due to water absorption effects.

\textbf{Salinas} is made up entirely of 16 different vegetation classes. The classes are larger with the smallest class comprising several hundred pixels. The scene has two large classes: "grapes untrained" and "vineyard untrained" which dominate a large area of land cover. We remove bands [108-112], [154-167] and 224 due to water absorption effects.

\textbf{University of Pavia} is different from the other benchmarks in that it is contains a significant amount of non-organic land cover such as asphalt and bricks. This scene contains a small number of classes and a more complex geometry which should make it harder for a superpixel based classifier to classify. 

\subsection{Description of the performance metrics}
In the main paper, we use three commonly used evaluation criteria to evaluate the performance of each classifier. In this section, we give the explicit description of each of these criteria.

\textbf{Overall Accuracy (OA).} This measure is the ratio of the number of correctly classified pixels divided by the total number of pixels. 

\textbf{Average Accuracy (AA).} This measure gives the average classification accuracy of all classes in an image. 

\textbf{Kappa Coefficient.} This metric gives the agreement between the final classification and the ground-truth. It gives the percentage agreement corrected by the chance that this agreement is due to chance alone and is thought to be more robust that simple percentage agreement.

\subsection{Parameter Analysis}

In the main paper, we explained how the classification accuracy changed with the parameter $K$. In this section we explain how the parameters $\sigma_l$ and $\beta$ change value depending on the HSI used. \medskip

$\sigma_l$  is the deviation of the location based Gaussian kernel $l_{ij}$.  Consider increasing the image size whilst keeping $\sigma_l$ constant. The width of the Gaussian distribution would become narrower and narrower with respect to the size of the image. This would reduce the weight of the edges connecting superpixels that are further apart compared to superpixel pairs which are close. Eventually the decreasing width of the kernel would lead to the removal of all non-local connections in the graph preventing information from properly propagating across the graph. Therefore, to balance the width of the spatial  kernel $l_{ij}$ to the spectral kernel $s_{ij}$, which does not change with image size, the value of $\sigma_l$ should increase.

$\beta$ weights between the two different spatial-spectral features $\vec{\mathcal{S}}^m_i$ and $\vec{\mathcal{S}}^w_i$ , with a lower value favouring the mean filter whilst a higher value emphasises the weighted filter. It was found that mean filtering was more effective for classifying Pavia University whilst weighted filtering was more effective when classifying Indian Pines and Salinas. An initial explanation for this is that the more complex land cover structure of the Pavia University scene means that the spatial information between superpixels is less informative than the spatial information within a superpixel.

\subsection{Further Experimental Results}
(E3) As an additional experiment, we classified each HSI using the same parameters as the main paper with $10$ labelled samples per class. From this, we produced a class by class accuracy breakdown. The results for this experiment are contained in Tables \ref{IP10}. From these tables, we see that the SGL method produced the highest classification accuracy for the majority of individual classes in each HSI with particular dominance in the Salinas and Indian Pines images. For the Indian Pines scene SGL produced clear accuracy improvements for classes 3 (Corn-mintill) and 10 (Soybean-mintill) in particular. Similarly, in the Salinas scene, SGL produced large improvements in the classification of classes 8 (Grapes untrained) and 15 (Vinyard untrained) as could be seen from the visual classificaiton maps in the main paper.

\clearpage

\clearpage

\begin{table*}[!t]

    \caption{OA(\%) AA(\%), Kappa and a class by class breakdown obtained by different classifiers with ten training samples per class. The best results are highlighted in green.}
    \label{IP10}
    \centering
    \begin{tabular}{|c|c|c|c|c|c|c|c|} \hline
    \multicolumn{8}{|c|}{{\cellcolor[HTML]{EFEFEF}\textsc{Indiana Pines}}}  \\\hline
    \cellcolor[HTML]{EFEFEF}\textsc{Class} & \cellcolor[HTML]{EFEFEF}\textsc{SGL}  & \cellcolor[HTML]{EFEFEF}\textsc{LCMR \cite{LCMR}}  & \cellcolor[HTML]{EFEFEF}\textsc{SC-MK \cite{SCMK}} &  \cellcolor[HTML]{EFEFEF}\textsc{EPF \cite{EPF}}  & \cellcolor[HTML]{EFEFEF}\textsc{LBP \cite{LBP}}  & \cellcolor[HTML]{EFEFEF}\textsc{IFRF \cite{IFRF}} & \cellcolor[HTML]{EFEFEF}\textsc{SVM \cite{melgani2004classification}}\\ \hline
    1 & 98.04 $\pm$ 0.65\% & 99.44 $\pm$ 1.17\% & 98.33 $\pm$ 1.43\% & 53.78 $\pm$ 28.30\% & \cellcolor{green}100.0 $\pm$ 0\%  & 53.51 $\pm$ 25.90\% & 20.53 $\pm$ 5.10\%\\
    2 & 76.06 $\pm$ 9.19\% & 75.99 $\pm$ 10.18\% & \cellcolor{green}78.82 $\pm$ 6.80\% & 55.88 $\pm$ 11.32\% & 70.94 $\pm$ 7.05\% & 70.35 $\pm$ 10.65\% & 43.14 $\pm$ 8.51\%\\
    3 & \cellcolor{green}84.27 $\pm$ 6.04\% & 70.78 $\pm$ 8.08\% & 77.65 $\pm$ 9.18\% & 60.19 $\pm$ 17.99\% & 70.28 $\pm$ 12.13\% & 67.99 $\pm$ 7.80\% & 39.15 $\pm$ 8.39\% \\
    4 & 96.67 $\pm$ 2.73\% & 93.88 $\pm$ 10,78\% & 86.39 $\pm$ 12.62\% & 34.57 $\pm$ 12.83\% & \cellcolor{green}97.93 $\pm$ 3.23\% & 79.37 $\pm$ 12.24\% & 21.25 $\pm$ 3.97\%\\
    5 & 92.30 $\pm$ 7.48\% & 90.32 $\pm$ 10.14\% & 82.33 $\pm$ 10.56\% & \cellcolor{green}93.39 $\pm$ 5.30\% & 82.92 $\pm$ 7.94\% & 79.40 $\pm$ 13.64\% & 59.34 $\pm$ 11.59\%\\
    6 & \cellcolor{green}98.71 $\pm$ 0.60\% & 91.40 $\pm$ 4.18\% & 89.54 $\pm$ 7.43\% & 86.32 $\pm$ 10.34\% & 90.36 $\pm$ 5.49\% & 93.63 $\pm$ 4.96\% & 83.29 $\pm$ 3.83\%\\
    7 & \cellcolor{green}100.0 $\pm$ 0.00\% & \cellcolor{green}100.0 $\pm$ 0.00\% & \cellcolor{green}100.0 $\pm$ 0.00\% & 70.92 $\pm$ 39.13\% & \cellcolor{green}100.0 $\pm$ 0.00\% & 39.65 $\pm$ 23.77\% & 24.85 $\pm$ 11.06\%\\
    8 & \cellcolor{green}100.0 $\pm$ 0.00\% & 99.68 $\pm$ 0.23\% & 97.09 $\pm$ 9.19\% & 98.41 $\pm$ 3.83\% & \cellcolor{green}100.0 $\pm$ 0.00\% & 99.97 $\pm$ 0.07\% & 93.12 $\pm$ 4.02\%\\
    9 & \cellcolor{green}100.0 $\pm$ 0.00\% & \cellcolor{green}100.0 $\pm$ 0.00\% & \cellcolor{green}100.0 $\pm$ 0.00\% & 59.44 $\pm$ 26.70\% & \cellcolor{green}100.0 $\pm$ 0.00\% & 28.81 $\pm$ 21.34\% & 12.59 $\pm$ 8.36\%\\
    10 & \cellcolor{green}88.94 $\pm$ 6.52\% & 76.46 $\pm$ 7.31\% & 71.32 $\pm$ 10.49\% & 61.19 $\pm$ 11.43\% & 79.90 $\pm$ 5.05\% & 75.95 $\pm$ 9.75\% & 36.93 $\pm$ 10.25\%\\
    11 & 91.04 $\pm$ 7.49\% & 71.40 $\pm$ 6.33\% & 69.22 $\pm$ 12.69\% & 81.05 $\pm$ 8.72\% & 73.78 $\pm$ 6.78\% & \cellcolor{green}93.81 $\pm$ 4.23\% & 61.50 $\pm$ 3.96\% \\
    12 & 90.05 $\pm$ 4.14\% & \cellcolor{green}90.50 $\pm$ 3.66\% & 78.47 $\pm$ 17.31\% & 44.31 $\pm$ 14.00\% & 70.58 $\pm$ 6.99\% & 74.08 $\pm$ 10.25\% & 28.20 $\pm$ 5.78\%\\
    13 & \cellcolor{green}99.56 $\pm$ 0.15\% & 99.33 $\pm$ 0.25\% & 99.90 $\pm$ 0.22\% & 98.34 $\pm$ 3.38\% & 98.31 $\pm$ 2.97\% & 75.32 $\pm$ 15.02\% & 80.12 $\pm$ 6.41\%\\
    14 & \cellcolor{green}100.0 $\pm$ 0.00\% & 98.18 $\pm$ 3.54\% & 88.14 $\pm$ 2.69\% & 95.15 $\pm$ 4.04\% & 91.32 $\pm$ 5.03\% & 98.31 $\pm$ 1.34\% & 88.22 $\pm$ 4.03\%\\
    15 & \cellcolor{green}97.69 $\pm$ 6.92\% & 91.62 $\pm$ 10.39\% & 90.96 $\pm$ 11.42\% & 64.91 $\pm$ 23.49\% & 90.59 $\pm$ 10.07\% & 77.14 $\pm$ 11.42\% & 39.10 $\pm$ 8.79\% \\
    16 & \cellcolor{green}100.0 $\pm$ 0.00\% & 98.67 $\pm$ 3.79\% & 97.59 $\pm$ 1.50\% & 84.46 $\pm$ 7.54\% & 98.43 $\pm$ 1.14\% & 92.62 $\pm$ 14.18\%  & 87.71 $\pm$ 20.71\% \\ \hline \hline
    OA & \cellcolor{green}90.89 $\pm$ 2.98\% & 82.74 $\pm$ 2.32\% & 79.91 $\pm$ 2.60\% & 68.95 $\pm$ 2.01\% & 80.52 $\pm$ 2.03\%  & 80.86 $\pm$ 3.76\% & 51.20 $\pm$ 3.92\% \\
    AA & \cellcolor{green}92.16 $\pm$ 6.77\% & 90.48 $\pm$ 1.56\% & 87.86 $\pm$ 1.53\% & 71.39 $\pm$ 3.49\% & 88.46 $\pm$ 1.29\% & 74.99 $\pm$ 3.16\% & 51.19 $\pm$ 3.22\%\\
    Kappa & \cellcolor{green}87.5 $\pm$ 3.33\% & 80.51 $\pm$ 2.59\% & 77.31 $\pm$ 2.93\% & 65.02 $\pm$ 2.25\% & 78.09 $\pm$ 2.23\% & 78.45 $\pm$ 4.16\% & 45.41 $\pm$ 4.11\% \\ \hline
    \noalign{\vskip 5mm}
    \end{tabular}

    \begin{tabular}{|c|c|c|c|c|c|c|c|} \hline
    \multicolumn{8}{|c|}{{\cellcolor[HTML]{EFEFEF}\textsc{Univeristy of Pavia}}}  \\\hline
     \cellcolor[HTML]{EFEFEF}\textsc{Class} & \cellcolor[HTML]{EFEFEF}\textsc{SGL}  & \cellcolor[HTML]{EFEFEF}\textsc{LCMR \cite{LCMR}}  & \cellcolor[HTML]{EFEFEF}\textsc{SC-MK \cite{SCMK}} & \cellcolor[HTML]{EFEFEF}\textsc{EPF \cite{EPF}}  & \cellcolor[HTML]{EFEFEF}\textsc{LBP \cite{LBP}}  & \cellcolor[HTML]{EFEFEF}\textsc{IFRF \cite{IFRF}} & \cellcolor[HTML]{EFEFEF}\textsc{SVM \cite{melgani2004classification}} \\ \hline
      1     & 86.64  $\pm$  4.39\% & 79.29  $\pm$  7.09\%  & 72.48  $\pm$  13.89\% & \cellcolor{green}94.80  $\pm$  4.52\%  & 59.64  $\pm$  5.07\% & 68.30  $\pm$  7.67\% & 94.09  $\pm$  5.44\%  \\
     2     & \cellcolor{green}95.87  $\pm$  3.17\% & 87.67  $\pm$  8.05\% & 80.05  $\pm$  8.03\% & 89.55  $\pm$  6.61\%  & 69.72  $\pm$  8.12\% & 94.90  $\pm$  2.19\% & 85.59  $\pm$  2.55\% \\
     3     & 85.37  $\pm$  10.53\% & \cellcolor{green}90.96  $\pm$  4.45\% & 76.84  $\pm$  9.10\% & 62.03  $\pm$  23.65\% & 79.52  $\pm$  7.30\% & 53.78  $\pm$  10.49\% & 42.74  $\pm$  13.46\% \\
     4     & 87.44  $\pm$  3.77\% & \cellcolor{green}95.10  $\pm$  3.40\% & 94.77  $\pm$  2.74\% & 57.08  $\pm$  11.72\% & 66.44  $\pm$  7.33\% & 66.44  $\pm$  22.53\% & 59.85  $\pm$  10.48\%\\
     5     & 95.84  $\pm$  2.91\% & 97.03  $\pm$  6.17\% & \cellcolor{green}99.66  $\pm$  0.08\% & 91.20  $\pm$  5.64\% & 89.91  $\pm$  12.78\% & 99.63  $\pm$  1.10\% & 93.69  $\pm$  5.78\%\\
     6     & \cellcolor{green}99.92  $\pm$  0.19\% & 95.37  $\pm$  2.36\% & 76.24  $\pm$  6.62\% & 49.32  $\pm$  13.91\% & 89.33  $\pm$  4.03\% & 82.47  $\pm$  9.46\% & 39.38  $\pm$  9.21\% \\
     7     & \cellcolor{green}96.59  $\pm$  1.10\% & 92.58  $\pm$  8.13\% & 76.06  $\pm$  14.92\% & 66.86  $\pm$  13.46\% & 89.15  $\pm$  8.91\% & 63.32  $\pm$  12.76\%  & 42.26  $\pm$  10.38\%\\
     8     & \cellcolor{green}94.03  $\pm$  5.63\% & 84.67  $\pm$  5.57\% & 79.79  $\pm$  3.85\% & 75.57  $\pm$  10.98\% & 80.78  $\pm$  16.55\% & 55.33  $\pm$  7.28\% & 73.22  $\pm$  5.74\% \\
     9     & 97.55  $\pm$  0.43\% & 93.80  $\pm$  3.55\% & \cellcolor{green}100.0  $\pm$  0.00\% & 98.48  $\pm$  1.70\% & 59.40  $\pm$  7.01\% & 49.33  $\pm$  9.07\% & 99.87  $\pm$  0.10\%\\ \hline \hline
     OA    & \cellcolor{green}93.70  $\pm$  1.35\% & 88.29  $\pm$  4.06\% & 80.23  $\pm$  4.06\% & 73.92  $\pm$  7.06\% & 72.66  $\pm$  4.29\% & 76.36  $\pm$  3.81\% & 67.40  $\pm$  4.66\% \\
     AA    & \cellcolor{green}93.25  $\pm$  5.03\% & 90.72  $\pm$  1.67\% & 83.99  $\pm$  2.15\% & 76.10  $\pm$  5.06\% & 75.99  $\pm$  2.71\% & 70.39  $\pm$  3.24\% & 70.08  $\pm$  2.48\% \\
     Kappa & \cellcolor{green}91.71  $\pm$  1.73\% & 84.91  $\pm$  4.89\% & 74.63  $\pm$  4.60\% & 67.40  $\pm$  8.23\% & 72.66  $\pm$  4.25\% & 69.70  $\pm$  4.55\% & 59.38  $\pm$  4.88\%\\ \hline
    \noalign{\vskip 5mm}    
    \end{tabular}

    \begin{tabular}{|c|c|c|c|c|c|c|c|} \hline
    \multicolumn{8}{|c|}{{\cellcolor[HTML]{EFEFEF}\textsc{Salinas}}}  \\\hline
    \cellcolor[HTML]{EFEFEF}\textsc{Class} & \cellcolor[HTML]{EFEFEF}\textsc{SGL} &\cellcolor[HTML]{EFEFEF}\textsc{ LCMR \cite{LCMR}} & \cellcolor[HTML]{EFEFEF}\textsc{SC-MK \cite{SCMK}} & \cellcolor[HTML]{EFEFEF}\textsc{EPF \cite{EPF}} & \cellcolor[HTML]{EFEFEF}\textsc{LBP \cite{LBP}} & \cellcolor[HTML]{EFEFEF}\textsc{IFRF \cite{IFRF}} & \cellcolor[HTML]{EFEFEF}\textsc{SVM \cite{melgani2004classification}}\\ \hline
    1 & \cellcolor{green}100.0 $\pm$ 0.00\%  & 99.95 $\pm$ 0.06\% & 99.93 $\pm$ 0.13\% & \cellcolor{green}100.0 $\pm$ 0.00\% & 97.97 $\pm$ 2.64\% & 95.77 $\pm$ 6.65\% & 97.54 $\pm$ 2.53\% \\
    2 & \cellcolor{green}100.0 $\pm$ 0.00\% & 93.21 $\pm$ 5.19\% & 98.66 $\pm$ 1.82\% & 99.87 $\pm$ 0.29\% & 96.57 $\pm$ 2.58\% & \cellcolor{green}100.0 $\pm$ 0.00\% & 99.10 $\pm$ 0.49\% \\
    3 & \cellcolor{green}100.0 $\pm$ 0.00\% & 99.56 $\pm$ 0.42\% & 96.94 $\pm$ 4.04\% & 93.84 $\pm$ 2.01\% & 98.59 $\pm$ 2.03\% & 99.32 $\pm$ 0.78\% & 86.62 $\pm$ 3.31\%\\
    4 & 99.71 $\pm$ 0.01\% & \cellcolor{green}100.0 $\pm$ 0.00\% & 98.79 $\pm$ 0.77\% & 97.70 $\pm$ 0.79\% & 97.84 $\pm$ 2.98\% & 87.42 $\pm$ 8.54\% & 96.92 $\pm$ 0.73\% \\
    5 & 98.09 $\pm$ 0.00\% & 96.88 $\pm$ 1.09\% & 95.63 $\pm$ 1.92\% & 99.48 $\pm$ 0.98\% & 92.37 $\pm$ 4.34\% & \cellcolor{green}99.92 $\pm$ 0.08\% & 97.57 $\pm$ 2.25\%\\
    6 & 99.93 $\pm$ 0.02\% & 98.53 $\pm$ 0.67\% & 99.53 $\pm$ 0.81\% & 99.98 $\pm$ 0.02\% & 92.14 $\pm$ 4.40\% & \cellcolor{green}100.0 $\pm$ 0.00\% & 99.97 $\pm$ 0.05\%\\
    7 & \cellcolor{green}99.48 $\pm$ 0.98\% & 97.57 $\pm$ 1.96\% & 94.22 $\pm$ 5.79\% & 97.92 $\pm$ 2.40\% & 92.68 $\pm$ 6.81\% & 98.88 $\pm$ 1.14\% & 97.68 $\pm$ 1.84\%\\
    8 & \cellcolor{green}99.38 $\pm$ 0.62\% & 87.84 $\pm$ 4.77\% & 74.47 $\pm$ 11.72\% & 84.19 $\pm$ 7.87\% & 85.27 $\pm$ 6.12\% & 96.83 $\pm$ 4.43\% & 70.82 $\pm$ 3.92\%\\
    9 & \cellcolor{green}100.0 $\pm$ 0.00\% & 96.90 $\pm$ 2.63\% & 99.40 $\pm$ 0.81\% & 99.47 $\pm$ 0.19\% & 93.05 $\pm$ 2.72\% & 98.82 $\pm$ 0.18\% & 98.84 $\pm$ 0.90\%\\
    10 & 96.72 $\pm$ 2.92\% & 93.71 $\pm$ 7.73\% & 88.34 $\pm$ 7.20\% & 86.13 $\pm$ 5.46\% & 93.65 $\pm$ 3.28\% & \cellcolor{green}99.21 $\pm$ 8.00\% & 79.64 $\pm$ 4.15\% \\
    11 & 95.882 $\pm$ 2.19\% & \cellcolor{green}99.94 $\pm$ 0.05\% & 97.03 $\pm$ 3.65\% & 91.81 $\pm$ 8.68\% & 97.83 $\pm$ 3.26\% & 98.96 $\pm$ 0.45\% & 83.29 $\pm$ 6.77\%\\
    12 & \cellcolor{green}99.90 $\pm$ 0.00\% & 99.60 $\pm$ 1.14\% & 97.50 $\pm$ 6.22\% & 99.42 $\pm$ 0.56\% & 89.96 $\pm$ 4.07\% & 98.19 $\pm$ 1.15\% & 94.42 $\pm$ 1.60\%\\
    13 & \cellcolor{green}98.80 $\pm$ 0.00\% & 98.65 $\pm$ 0.73\% & 95.36 $\pm$ 4.40\% & 96.32 $\pm$ 2.87\% & 91.59 $\pm$ 6.07\% & 92.20 $\pm$ 8.00\% & 88.15 $\pm$ 8.68\%\\
    14 & \cellcolor{green}95.38 $\pm$ 1.48\% & 95.23 $\pm$ 2.86\% & 90.29 $\pm$ 6.73 & 95.27 $\pm$ 11.66\% & 88.18 $\pm$ 6.84\% & 87.05 $\pm$ 14.28\% & 84.51 $\pm$ 17.07\%\\
    15 & \cellcolor{green}99.28 $\pm$ 0.07\% & 88.12 $\pm$ 7.50\% & 84.36 $\pm$ 6.15\% & 56.02 $\pm$ 5.52\% & 82.42 $\pm$ 12.45\% & 86.51 $\pm$ 8.12\% & 49.19 $\pm$ 2.71\%\\
    16 & \cellcolor{green}100.0 $\pm$ 0.00\% & 94.46 $\pm$ 6.47\% & 96.17 $\pm$ 3.18\% & 98.67 $\pm$ 4.09\% & 97.96 $\pm$ 3.91\% & 99.77 $\pm$ 0.56\% & 92.51 $\pm$ 8.59\%\\ \hline \hline
    OA & \cellcolor{green}99.24 $\pm$ 0.16\% & 93.90 $\pm$ 1.29\% & 90.38 $\pm$ 2.42\% & 86.53 $\pm$ 1.99\% & 90.68 $\pm$ 1.35\% & 95.87 $\pm$ 1.62\% & 82.42 $\pm$ 1.15\% \\
    AA & \cellcolor{green}98.9 $\pm$ 1.51\% & 96.26 $\pm$ 1.02\% & 94.16 $\pm$ 1.11\% & 93.51 $\pm$ 0.91\% & 93.00 $\pm$ 1.03\% & 96.24 $\pm$ 1.43\% & 88.55 $\pm$ 0.99\% \\
    Kappa & \cellcolor{green} 99.15 $\pm$ 0.17\% & 93.22 $\pm$ 1.44\% & 89.33 $\pm$ 2.663\% & 85.10 $\pm$ 2.15\% & 90.68 $\pm$ 1.36\% & 95.41 $\pm$ 1.80\% & 80.53 $\pm$ 1.25\%\\ \hline 

    \end{tabular}

\end{table*}

\end{document}